%% file: arxiv.tex
\documentclass{article}
\usepackage{arxiv}
\usepackage{fullpage}
\usepackage{charter}

\input{math_commands.tex}

\usepackage[numbers]{natbib}
\usepackage[hypertexnames=false]{hyperref}
\usepackage{url}
\usepackage{float}
\usepackage{graphicx}
\usepackage{booktabs}
\usepackage{mathtools}
\usepackage{enumitem}
\setlist{nosep}

\floatstyle{ruled}
\newfloat{algorithm}{tbp}{loa}
\floatname{algorithm}{Algorithm}
\newenvironment{algorithmic}
  {\begin{center}\begin{minipage}{0.96\linewidth}\setlength{\parindent}{0pt}\setlength{\parskip}{2pt}}
  {\end{minipage}\end{center}}
\newcommand{\algline}[1]{\par\noindent #1}
\newcommand{\algindentline}[1]{\par\noindent\hspace*{1.5em}#1}
\newcommand{\algdeepindentline}[1]{\par\noindent\hspace*{3em}#1}
\newcommand{\indicator}{\mathbf{1}}
\newcommand{\Return}{\textbf{return} }

\title{Large Language Model Selection\\ with Limited Annotations}

\author{
Yavuz Durmazkeser \\
TU Delft \\
\texttt{y.durmazkeser@tudelft.nl} \\ \\ \\
\And
Patrik Okanovic \\
ETH Zurich \\
\texttt{patrik.okanovic@inf.ethz.ch} \\ \\ \\
\And
Andreas Kirsch \\
\texttt{blackhc@gmail.com} \\
\And
Torsten Hoefler \\
ETH Zurich \\
\texttt{htor@ethz.ch} \\
\And
Nezihe Merve G\"urel \\
TU Delft \\
\texttt{n.m.gurel@tudelft.nl} \\
}

\begin{document}

\maketitle

\begin{abstract}
Choosing a Large Language Model (LLM) for a given task requires comparing many strong candidates, yet standard evaluation relies on costly annotations over fixed evaluation sets. To address this challenge, we develop \framework, the first framework for active model selection of LLMs. \framework{} aims to find a small set of queries whose annotations are most informative for identifying the best LLM for a given task. To this end, we introduce a query selection rule based on expected information gain, computed from pairwise similarities between candidate model outputs. Because this rule only uses generated model responses, \framework{} can be applied across candidate models without assumptions about their architecture or access to model weights. This makes it suitable for both open-weight and black-box LLMs. We evaluate \framework{} across 23 datasets, 156 evaluated models, diverse task families, and multiple text evaluation metrics. Across all experiments, \framework{} improves over the strongest baseline in every setting, with annotation cost reductions up to 81.8\% for best model selection and up to 84.78\% for near-best model selection.
\end{abstract}

\section{Introduction}
The strong zero-shot abilities of modern Large Language Models (LLMs) allow them to be deployed directly across many tasks and domains without task-specific retraining or fine-tuning \cite{weifinetuned, NEURIPS2022_8bb0d291}. This creates a natural model selection problem: given a target application or data distribution, one must choose which LLM to use. However, this choice has become increasingly difficult as the number of strong candidate models continues to grow \cite{openai2026gpt55, anthropic2026claudeopus47, google2026gemini31pro}. Academic repositories and commercial platforms now provide access to a large and diverse set of pretrained models that vary in architecture, training data, and optimization objectives \cite{huggingfacehub, openaiapi, geminiapi, anthropicapi}. As a result, identifying the best LLM for a particular task or domain has become an important challenge.

The key difficulty is that model quality is highly context-dependent: relative performance differs substantially across domains, tasks, and languages \cite{liang2023holistic}. Although evaluating and benchmarking text generation systems has long been an active area of research \cite{liang2023holistic, open-llm-leaderboard-v2, 2023opencompass}, the rapid growth of both available LLMs and possible evaluation settings makes standard benchmarking increasingly hard to use for model selection \cite{10.1145/3641289}. Existing benchmarks may lag behind the release of new models, and they often emphasize standardized or narrow tasks that do not fully reflect the needs of domain-specific applications. A natural alternative is to perform model selection using small annotated subsets of queries, chosen at random or through simple heuristics, but such subsets can use the annotation budget inefficiently and may fail to reveal meaningful differences among models \cite{zouhar2025select, pmlr-v139-kossen21a}. Active model selection addresses this issue by allocating limited annotations to strategically selected examples \cite{karimi2021online, 9101367, pmlr-v258-okanovic25a, NIPS2015_d9731321, tahan2024label}, but existing work has primarily focused on classification rather than generative settings \cite{pmlr-v258-okanovic25a, kay2025coda, madani2012activemodelselection, karimi2021online, piratla2021active, liu2022contextual, kassraie2023anytime, xia2024convergenceaware, li2024necessity, li2024online}. Consequently, how to reliably identify the best LLM for a target task and data distribution under limited annotation resources remains an open question.

In this work, we study active model selection for LLMs under limited annotation budgets. Given a query pool and a set of candidate LLMs, our goal is to determine which queries should be annotated so that the best model can be identified reliably, while using as few annotations as possible. Our contributions are summarized below:
\begin{itemize}
    \setlength{\itemsep}{2pt}
    \setlength{\parskip}{0pt}
    \setlength{\parsep}{0pt}
    \setlength{\topsep}{2pt}
    \item We introduce, to the best of our knowledge, the first formulation and study of active model selection for LLMs under limited annotation budgets.
    \item We propose \framework{}, a principled and model-agnostic framework that greedily selects $b$ queries from a large set of $n$ queries, where $b \ll n$, whose annotations are expected to maximally reduce uncertainty about the best model for the entire set. Motivated by near-optimality results for greedy sequential information maximization \cite{pmlr-v40-Chen15b}, our approach builds on information gain criteria and derives a tractable approximation that reduces query selection to a posterior-weighted function of pairwise similarities between generated model responses.
    \item We validate \framework\ against adapted baseline strategies in a reference-based evaluation setting across $23$ datasets and 156 evaluated models, covering diverse task families and standard task-specific metrics, including BERTScore \cite{zhang2020bertscoreevaluatingtextgeneration}, ROUGE \cite{lin-2004-rouge}, BLEU \cite{papineni-etal-2002-bleu}, F1 score, mathematical equivalence \cite{hendrycks2021measuring}, and exact match. \framework\ identifies the best model using much fewer annotations than the strongest baseline across all $23$ datasets, with reductions of up to $81.8\%$. This advantage also carries over to near-best model selection, where reductions reach $84.78\%$, and to robustness evaluations based on the worst-case performance gap between the selected model and the best model.
\end{itemize}

\begin{figure}
  \centering
  \includegraphics[width=\textwidth]{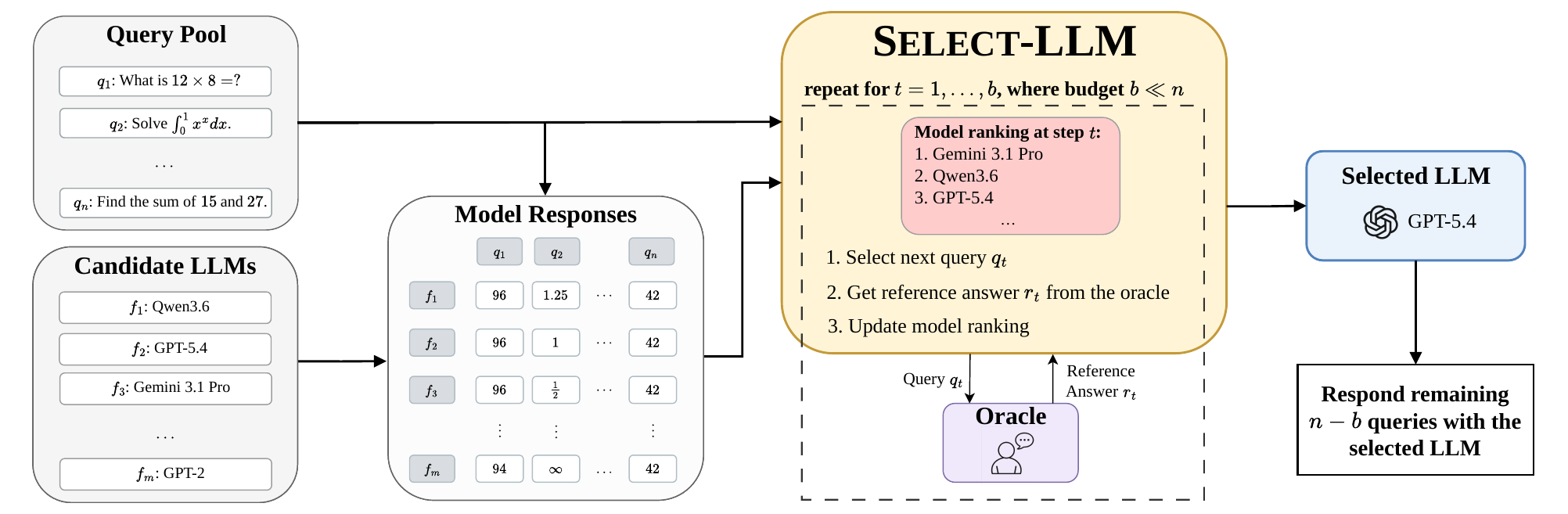}
  \vspace{-5pt}
  \caption{An overview of \framework. For an arbitrary pool of $n$ queries and a set of candidate language models, \framework {} adaptively annotates most informative $b \ll n$ queries for identifying the best language model for the pool.}
  \vspace{-5pt}
  \label{fig:figure1}
\end{figure}

After \framework{} selects an LLM based on $b$ annotated queries, the selected model generates outputs for the remaining $n-b$ queries, where $n-b \gg b$. \framework{} uses only model responses, requires no access to parameters, logits, or internal states, and imposes no restrictions on output format, making it applicable to black-box and API-only LLMs. Figure \ref{fig:figure1} gives an overview of the full selection procedure.

\section{Related Work}
Several paradigms exist for \textbf{LLM evaluation}. Standard benchmarks use fixed multiple-choice \cite{srivastava2022beyond, suzgun2022challenging} or short-answer tasks \cite{cobbe2021gsm8k}, while open-ended tasks such as summarization \cite{46111, narayan-etal-2018-dont} and translation \cite{goyal-etal-2022-flores} are often evaluated against human references using BLEU \cite{papineni-etal-2002-bleu}, ROUGE \cite{lin-2004-rouge}, or BERTScore \cite{zhang2020bertscoreevaluatingtextgeneration}. More recent judge-based evaluations use human preferences, as in LMArena \cite{10.5555/3666122.3668142}, or LLM-as-a-Judge, as in Arena-Hard \cite{arenahard2024}, AlpacaEval \cite{alpaca_eval}, and MT-Bench \cite{10.5555/3666122.3668142}. These methods are primarily designed for full-benchmark assessment on predefined evaluation sets. In contrast, our setting starts from a task-specific pool of queries, possibly newly collected for the application, and asks which queries should be annotated to select the best model under a limited annotation budget.

Recent work on \textbf{benchmark and ranking approximation} reduces evaluation costs by replacing full evaluations with smaller evaluation sets or cheaper evaluation procedures. Mini-benchmarking methods summarize existing benchmarks by selecting compact subsets that preserve full-benchmark scores or rankings \cite{vivek-etal-2024-anchor, 10.5555/3692070.3693466, li-xiong-2025-towards, saranathan-etal-2025-sublime}. Related work studies efficient benchmark construction \cite{perlitz2024efficient}, ranking-preserving subset selection \cite{zouhar2025select}, and efficient pairwise comparison protocols for ranking LLMs \cite{feng2025sample}. These methods are best viewed as evaluation summarization approaches: they aim to approximate full-benchmark scores or model rankings using fewer examples, generations, or pairwise comparisons. By contrast, active model selection under limited annotations begins before oracle annotations are available and must decide which target examples to annotate. Accordingly, our goal is not to approximate full-benchmark evaluation from existing annotations, but to identify the best model for a given task or domain within the given limited budget. 

\textbf{Active model selection} has largely been developed outside the LLM setting. Prior work primarily considers classification problems \cite{4518001, 9101367, NIPS2015_d9731321, pmlr-v258-okanovic25a, kay2025coda}, with some studies focusing on an online setting where examples arrive sequentially from a stream \cite{madani2012activemodelselection, karimi2021online, piratla2021active, liu2022contextual, kassraie2023anytime, xia2024convergenceaware, li2024necessity, li2024online, xia2024llm, poon2026multillm}. Existing LLM-oriented methods are more limited in scope, either restricting the comparison to two candidate models \cite{tahan2024label} or focusing on active testing of a single model \cite{berrada2025scaling, huang2025actracer, li2025activeevaluation}. As a result, they do not address selection from arbitrary-size pools of candidate LLMs. \framework{} fills this gap by selecting among any number of LLMs under a limited annotation budget, using only model responses and requiring no access to internal states or logits.

\section{\framework{}}
\label{sec:llm-selector}
In this section, we introduce \framework. We first define the problem setting in Section~\ref{subsec:problem_setting}. In Section~\ref{subsec:algorithm}, we present our model-agnostic sequential query selection algorithm for annotation-efficient LLM selection. Section~\ref{subsec:parameter_selection} describes our hyperparameter selection strategy, which requires no oracle annotations.

\subsection{Problem Setting}
\label{subsec:problem_setting}
Consider the inference-time scenario in which we are provided with a set of $n$ unannotated queries $Q = \{ q_i \in \mathcal{Q} \mid i \in [n] \}$.
Each query $q_i$ represents a user-issued prompt or request to an oracle.  
We denote the oracle-annotated ground-truth response to $q_i$ by $r_i \in \mathcal{R}$.  
Since these annotations are not observed, we use $R_i$ to denote the corresponding unknown random variable.

Given a collection of $m$ pretrained language models $\mathcal{M} = \{ f_j : \mathcal{Q} \to \mathcal{R} \mid j \in [m] \}$, our objective is to identify the best language model in $\mathcal{M}$ for producing high-quality responses to the queries $Q$.  
Because oracle-provided annotations are costly, we assume access to only a limited number of at most $b \ll n$ annotations.  
The problem therefore reduces to selecting $b$ queries whose annotations provide maximal information about the identity of the best model.  

We define the best model, denoted by $f^*$, as the model that achieves the highest performance over the query set $Q$ when evaluated against ground-truth responses.
Concretely, for each model $f_j$, performance is measured using standard reference-based evaluation metrics (e.g., ROUGE \cite{lin-2004-rouge}, BLEU \cite{papineni-etal-2002-bleu}, BERTScore \cite{zhang2020bertscoreevaluatingtextgeneration}, F1 score, or exact match) that compare model outputs $f_j(q_i)$ with the reference answers $r_i$.  
The best model $f^*$ is then the one with the highest average score across all queries.

Formally, we cast the selection problem as one of maximizing mutual information.  
We introduce a random variable $F$ taking values in $\mathcal{M}$, which represents the unknown identity of the best model.  
Our goal is to reduce uncertainty about $F$ by selectively acquiring annotations.
That is, we aim to identify a subset $\mathcal{A} \subseteq \{(q_i, r_i) \mid i \in [n]\}$ with $|\mathcal{A}| \leq b$ whose annotations maximize the mutual information with $F$:
\begin{equation}
A_{\mathrm{opt}[b]}
    = \argmax_{\substack{\mathcal{A} \subseteq \{(q_i, r_i) \mid i \in [n]\} \\ |\mathcal{A}| \leq b}}
      \mathbb{I}(F; \mathcal{A}).\label{Eq: objective}
\end{equation}

To formalize the behavior of the best model, we adopt a probabilistic formulation. For a given query $q \in \mathcal{Q}$, reference response $r \in \mathcal{R}$, and similarity function $s : \mathcal{R} \times \mathcal{R} \to \mathbb{R}$ (e.g., ROUGE, BLEU, or BERTScore), we take $1-s(r,F(q))$ as the distance between the reference response and the model output. We then use this distance in a Laplace-type distribution over the response space, so that outputs more similar to the reference response receive higher probability. In particular, the model assigns probability proportional to $\exp(-(1-s(r,F(q)))/\tau)$, or equivalently $\exp(s(r,F(q))/\tau)$ up to a constant factor absorbed into the normalizer $Z(\tau)$. This gives
\begin{equation}
\mathbb{P}\big(F(q) = r \mid F = f^*, q\big)
= \frac{1}{Z(\tau)} \exp\!\left(\frac{s(r, F(q))}{\tau}\right),
\label{eq:score-based-likelihood}
\end{equation}
where $\tau \in (0, \infty)$ is a temperature hyperparameter controlling the sharpness of the distribution. We use this Laplace modeling as it guarantees that the average likelihood over all queries in $Q$ induces the same candidate model ranking as the average score.

This formulation assigns higher probability to responses that are more similar to the reference answer, while $\tau$ governs how strongly the distribution concentrates on high-scoring responses. The temperature parameter $\tau$ is selected before active query selection and does not introduce additional oracle annotation cost. Details of this procedure are provided in Section~\ref{subsec:parameter_selection}.

\subsection{The Algorithm}
\label{subsec:algorithm}
Given the query set $Q$, our objective is to select at most $b$ queries such that, once annotated, they maximize our information about the best language model as defined in~\Eqref{Eq: objective}. To this end, we adopt a sequential information maximization strategy~\cite{pmlr-v40-Chen15b, pmlr-v258-okanovic25a} for selecting queries one at a time until the budget $b$ is exhausted. 

In our sequential framework, let $U_t$ denote the pool of unannotated queries, and $A_t$ the set of annotated queries accumulated up to sequential step $t$, with $U_0 = Q$ and $A_0 = \emptyset$. At each step $t$, we select the next query $q_t$ according to
\begin{equation}
q_t = \argmax_{q \in U_t} \quad \mathbb{I}(F;R \mid A_t, q).
\label{eq: query selection}
\end{equation}

Expanding the conditional mutual information, using the conditional independence of $R$ from $A_t$ given $(F,q)$, and noting that the candidate query $q$ is fixed when computing the score, we obtain
\begin{align}
\mathbb{I}(F;R \mid A_t,q)
&= D_{\mathrm{KL}}\!\left(
\mathbb{P}(F,R \mid A_t,q)
\;\middle\|\;
\mathbb{P}(F \mid A_t,q)\mathbb{P}(R \mid A_t,q)
\right) \nonumber \\
&= \sum_{j=1}^m \sum_{r \in \mathcal{R}}
\mathbb{P}(F=f_j \mid A_t)\mathbb{P}(R=r \mid F=f_j,q)
\log
\frac{\mathbb{P}(R=r \mid F=f_j,q)}
{\mathbb{P}(R=r \mid A_t,q)}.
\label{eq: mi-expanded}
\end{align}

Here, $D_{\mathrm{KL}}(\cdot \| \cdot)$ denotes the Kullback-Leibler (KL) divergence. The posterior $\mathbb{P}(F = f_j \mid A_t)$ is computed using the likelihood formulation defined in Section~\ref{subsec:problem_setting} together with Bayes' rule. In particular, given a set of annotated pairs $A_t = \{(q_\ell, r_\ell)\}_{\ell=1}^t$, the posterior takes the form $\mathbb{P}(F = f_j \mid A_t)
\propto
\mathbb{P}(F = f_j) \prod_{\ell=1}^t
\exp\!\left(\frac{s(r_\ell, f_j(q_\ell))}{\tau}\right).$
We use a uniform prior over the candidate best model, assigning equal prior probability to each $f_j \in \mathcal{M}$. This choice is made for simplicity; when prior information about the candidate models is available, it can be incorporated through a non-uniform prior.

Directly computing~\Eqref{eq: mi-expanded} is intractable for open-ended generation, because the summation is over the unbounded reference space $\mathcal{R}$. We therefore derive a surrogate that preserves the information gain principle while replacing the summation over possible references with a comparison among the candidate model outputs. The full derivation of the final selection rule and an analysis of the approximation are provided in Appendix~\ref{sec:appendix-derivation}. The approximation has three main steps:
\begin{itemize}
    \setlength{\itemsep}{2pt}
    \setlength{\parskip}{0pt}
    \setlength{\parsep}{0pt}
    \setlength{\topsep}{2pt}
    \item We approximate the logarithmic ratio in \Eqref{eq: mi-expanded} by applying a second-order Taylor approximation of $x \log x$ around $x=1$. This yields a quadratic approximation to the information gain, which can be rewritten as a pairwise comparison involving
    $\left(
    \frac{\mathbb{P}(R = r \mid F = f_j, q)}{\mathbb{P}(R = r \mid A_t, q)}
    -
    \frac{\mathbb{P}(R = r \mid F = f_k, q)}{\mathbb{P}(R = r \mid A_t, q)}
    \right)^2$,
    averaged over possible reference responses $r \in \mathcal{R}$.
    \item We locally linearize the likelihood model in \Eqref{eq:score-based-likelihood}. In particular,
    $\mathbb{P}(R = r \mid F = f_j, q)-\mathbb{P}(R = r \mid F = f_k, q)$
    is approximated by
    $\frac{1}{Z(\tau)\tau}\big(s(r,f_j(q))-s(r,f_k(q))\big)$,
    turning the average over likelihood differences into an average over similarity differences between each model output and the same possible reference response.
    \item This average over possible reference responses is then approximated by a quantity proportional to the dissimilarity between the two model outputs, measured by $1-s(f_j(q),f_k(q))$.
\end{itemize}

After these approximations, all remaining proportionality constants are independent of $q$. Dropping these constants and reversing the sign converts the maximization of information gain into an equivalent minimization problem. This leads to the tractable selection rule:
\begin{equation}
q_t
=
\argmin_{q \in U_t}
\sum_{j=1}^m \sum_{k=1}^m
\mathbb{P}(F = f_j \mid A_t)\mathbb{P}(F = f_k \mid A_t)
s(f_j(q), f_k(q)).
\label{eq: final-query-selection-rule}
\end{equation}

This rule follows from the information gain objective and has a natural disagreement interpretation: if two plausible best models give similar responses, the query is unlikely to separate them; if they give dissimilar responses, the query is more informative. Thus, the acquisition rule prioritizes queries whose responses are most likely to distinguish between competing candidates for the best model.

After obtaining the annotation $r_t$ for the selected query $q_t$, we update the posterior over the best model using Bayes' rule. In particular, for each model $f_j \in \mathcal{M}$, the posterior is updated as
\begin{equation}
\mathbb{P}(F = f_j \mid A_{t+1})
\propto
\mathbb{P}(F = f_j \mid A_t)\,\mathbb{P}(R = r_t \mid F = f_j, q_t),
\end{equation}
where $A_{t+1} = A_t \cup \{(q_t, r_t)\}$. Using the likelihood model defined in Section~\ref{subsec:problem_setting}, this update takes the form
\begin{equation}
\mathbb{P}(F = f_j \mid A_{t+1})
\propto
\mathbb{P}(F = f_j \mid A_t)\exp\!\left(\frac{s(r_t, f_j(q_t))}{\tau}\right).
\end{equation}

In Algorithm~\ref{alg:llm-selector}, we present the full procedure of the proposed active model selection method.

\begin{algorithm}[H]
    \caption{\framework{} Algorithm}\label{alg:llm-selector}
    \small
    \begin{algorithmic}
    \algline{\textbf{Input:} model set $\mathcal{M} = \{f_j\}_{j=1}^m$, query set $Q$, similarity metric $s(\cdot,\cdot)$, temperature parameter $\tau$, labeling budget $b$}
    \algline{$A_0 \gets \emptyset$, $U_0 \gets Q$}
    \algline{\textbf{for} $j = 1$ to $m$ \textbf{do}}
    \algindentline{$\mathbb{P}(F = f_j \mid A_0) \gets \frac{1}{m}$}
    \algline{\textbf{end for}}
    \algline{\textbf{for} $t = 0$ to $b-1$ \textbf{do}}
    \algindentline{$q_t \gets \displaystyle\argmin_{q \in U_t}
    \sum_{j=1}^m \sum_{k=1}^m
    \mathbb{P}(F = f_j \mid A_t)\mathbb{P}(F = f_k \mid A_t)
    s(f_j(q), f_k(q))$}
    \algindentline{$r_t \gets \operatorname{Oracle}(q_t)$}
    \algindentline{$A_{t+1} \gets A_t \cup \{(q_t, r_t)\}$}
    \algindentline{$U_{t+1} \gets U_t \setminus \{q_t\}$}
    \algindentline{\textbf{for} $j = 1$ to $m$ \textbf{do}}
    \algdeepindentline{$\mathbb{P}(F = f_j \mid A_{t+1})
    \gets
    \frac{1}{Z}
    \mathbb{P}(F = f_j \mid A_t)
    \exp\!\left(\frac{s(r_t, f_j(q_t))}{\tau}\right)$}
    \algindentline{\textbf{end for}}
    \algline{\textbf{end for}}
    \algline{\Return $\argmax_{f_j \in \mathcal{M}} \mathbb{P}(F = f_j \mid A_b)$}
    \end{algorithmic}
\end{algorithm}

\noindent\textbf{Parameter Selection.}
\label{subsec:parameter_selection}
We optimize the temperature parameter $\tau$ prior to the model selection procedure, without requiring any oracle annotations. We follow a procedure similar to~\cite{pmlr-v258-okanovic25a}. Specifically, we construct a noisy proxy for the reference similarity score by computing, for each query, the average similarity between a model's response and the responses of all models in $\mathcal{M}$. Formally, for model $f_j$ and query $q_i$, we define the noisy proxy score as
$\tilde{s}_{i,j}
=
\frac{1}{m}
\sum_{k=1}^m
s\bigl(f_j(q_i), f_k(q_i)\bigr).$

Using this proxy, we simulate the model selection procedure under noisy annotations and evaluate its performance for different values of $\tau$. We then perform a grid search over $\tau$ and select the value that yields the highest performance under the noisy annotations.

\section{Experiments}
\label{sec:experiments}

In this section, we evaluate \framework\ on a diverse collection of LLM benchmarks and compare it against adapted active model selection baselines. We first describe the dataset and model collections in Section~\ref{subsec:dataset_model_collections} and the baselines in Section~\ref{sec:baselines}. We then present the experimental setup in Section~\ref{sec:experimental_setup} and report results on best model identification, near-best model selection, and robustness in Section~\ref{sec:results}.

\subsection{Dataset and Model Collections}
\label{subsec:dataset_model_collections}
We evaluate on $23$ datasets covering reasoning, graduate-level question answering, language understanding, summarization, image captioning, mathematics, sentiment classification, reading comprehension and open-domain question answering, visual question answering, and machine translation. The benchmark suite also includes setup variants when available, such as Chain-of-Thought prompting, closed-book and open-book question answering, and multiple metrics. All experiments use precomputed model outputs from HELM, VHELM, HELM Lite, and Open LLM Leaderboard v2 \cite{liang2023holistic,lee2024vhelm,liang2023helmlite,open-llm-leaderboard-v2}; and we do not run additional LLM inference. These sources yield collections of LLMs that include both open-source and API-only models. We evaluate these models with task-specific reference-based metrics, including exact match, mathematical equivalence \cite{hendrycks2021measuring}, F1 score, BLEU-4 \cite{papineni-etal-2002-bleu}, ROUGE-L \cite{lin-2004-rouge}, and BERTScore \cite{zhang2020bertscoreevaluatingtextgeneration}. Appendix~\ref{sec:appendix-datasets-model-collections} provides additional details on the datasets, model collections, numbers of evaluated LLMs, dataset sizes, realization sizes, and evaluation metrics.

\subsection{Baselines}
\label{sec:baselines}

There are currently no standard baselines tailored to active model selection for an arbitrary number of LLMs. Prior active model selection methods largely focus on classification settings \cite{pmlr-v258-okanovic25a, kay2025coda}, while existing LLM-oriented work is restricted by the number of evaluated models \cite{tahan2024label, berrada2025scaling, huang2025actracer, li2025activeevaluation}. We therefore compare \framework\ against a set of representative strategies adapted to our setting, where selection must be made from model responses alone. Our baseline suite follows the active model selection comparison of~\cite{pmlr-v258-okanovic25a}, which adapts query-by-committee methods, \textsc{VMA}, and \textsc{AMC} for limited annotation model selection. Here, we further adapt these strategies to generative LLM outputs.

\begin{itemize}
    \setlength{\itemsep}{2pt}
    \setlength{\parskip}{0pt}
    \setlength{\parsep}{0pt}
    \setlength{\topsep}{2pt}
        \item \textbf{\textsc{Random}} samples queries uniformly at random from the unlabeled pool and serves as a non-adaptive reference baseline.
    \item \textbf{\textsc{Margin}} adapts query-by-committee style margin sampling \cite{seung1992query, freund1997selective}. For each query, we compute model support scores from average pairwise similarities between responses and prioritize queries with the smallest gap between the two highest support values.
    \item \textbf{\textsc{Min Agreement}} adapts committee-based uncertainty sampling \cite{dagan1995committee} to our setting by selecting queries whose maximum agreement score across candidate models is smallest.
    \item \textbf{\textsc{VMA}} follows the variance minimization strategy of \cite{matsuura2023active} and samples queries that reduce the variance of the estimated model risks. Since our setting uses only model responses, we adapt this method using risk estimates derived from pairwise similarities between model outputs.
    \item \textbf{\textsc{AMC}} adapts active model comparison \cite{NIPS2012_92fb0c6d}, originally designed for choosing between competing predictors under a labeling budget, to the multi-LLM setting by constructing model risks from pairwise similarities between responses and sampling queries that better distinguish competing models.
\end{itemize}

None of these methods was originally proposed for selecting the best LLM from a candidate set of generative models under limited annotation budgets. They nevertheless provide strong and diverse baselines for evaluating \framework\ in this newly formalized setting.

\subsection{Experimental Setup}
\label{sec:experimental_setup}

We adapt the evaluation protocol of \cite{pmlr-v258-okanovic25a} to the LLM setting. For each dataset, we form a realization by uniformly sampling $n$ evaluation instances without replacement from the full benchmark pool. We choose $n$ based on the number of available evaluation instances in each dataset, using $200$ for the two MATH datasets, $300$ for MuSR, $1000$ for BBH, MMLU-Pro, NarrativeQA, and Natural Questions, and $500$ for the remaining datasets. Each method then sequentially selects queries up to a budget $b$, with $b=n$ in all experiments. After each step $t \in \{1,\dots,b\}$, the currently selected model is the candidate with the highest average oracle score on the $t$ selected queries. The true best model for that realization is defined using all $n$ sampled instances. We repeat this process for $1000$ realizations per dataset and report averages across realizations.

For \framework, the temperature parameter $\tau$ is tuned separately for each dataset using the noisy annotation procedure described in Section~\ref{subsec:parameter_selection}. Concretely, we build a proxy signal from pairwise similarities between model responses, run the same realization-based evaluation on this proxy signal, and choose the value of $\tau$ that yields the strongest identification performance over a fixed grid, without using the oracle scores from the final evaluation runs. Additional details on parameter selection are provided in Appendix~\ref{sec:appendix-temperature-selection}.

We evaluate model selection methods using three metrics. \textit{Identification probability} is the fraction of realizations in which a method identifies the true best model at budget $t$. \textit{Annotation efficiency} measures the relative reduction in labels required by \framework\ to identify either the true best model or a near-best model. For the latter, we define a selected model as near-best if its average score is within a relative gap $\delta$ of the best model, that is, if its score is at least $(1-\delta)$ times the best score on that realization. \textit{95th percentile performance gap} is the $95$th percentile, across realizations, of the relative difference between the score of the true best model and that of the selected model.

\subsection{Results}
\label{sec:results}

\subsubsection{Best Model Identification Probability}

\begin{figure}[!htbp]
  \centering
  \includegraphics[width=\linewidth]{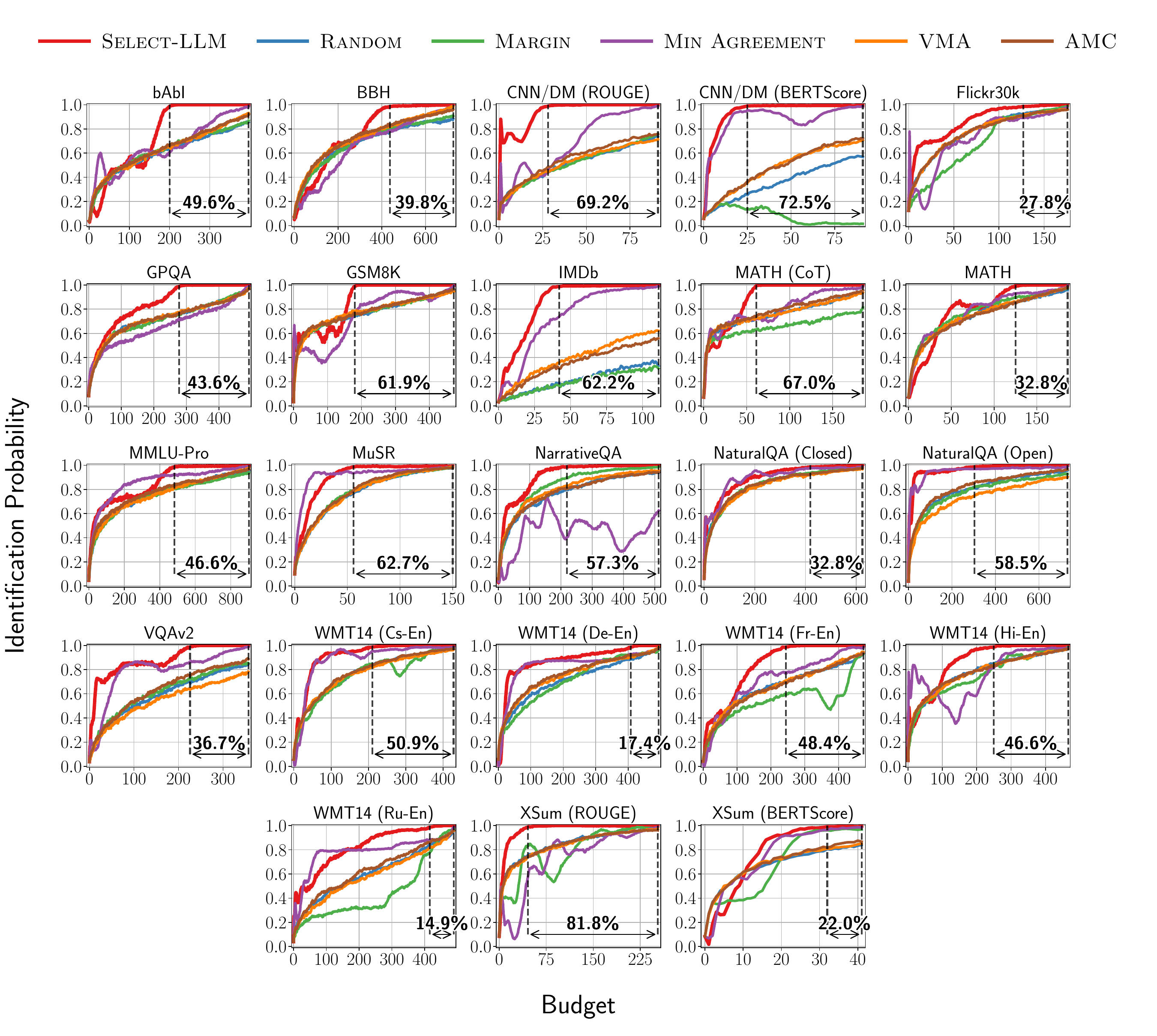}
  \vspace{-15pt}
  \caption{Comparison of \framework\ and the baselines in terms of best model identification probability across 23 datasets. In each plot, the horizontal arrow and percentage indicate \framework{}'s labeling efficiency relative to the strongest baseline, and the dashed vertical lines mark the corresponding budgets. Each plot is shown until the strongest baseline reaches $100\%$ identification probability.}
  \vspace{-5pt}
  \label{fig:identification_prob_results}
\end{figure}

Figure~\ref{fig:identification_prob_results} compares the identification probability curves of \framework\ and the baselines across all $23$ datasets. Across datasets, \framework\ identifies the best model using fewer labels than the strongest baseline; the savings are largest on XSum (ROUGE-L) at $81.8\%$ and remain positive even in the smallest case, WMT14 (Russian-English), at $14.9\%$. Among the baselines, \textsc{Min Agreement} is the strongest competitor on $17$ datasets, while \textsc{Margin} and \textsc{VMA} are the strongest baselines on $3$ datasets each.

The curves also show how the methods behave across the budget. On $6$ of the $23$ datasets, \textsc{Min Agreement} has an early lead over \framework{}, but this lead lasts at most the first $16\%$ of the budget. This behavior can be explained by the exploration phase of \framework, where the algorithm selects queries with the lowest overall agreement. The early identification probability of \framework\ therefore depends on whether the true best model succeeds on these exploratory queries. As more annotations are collected, \framework\ can better identify which queries are most informative for distinguishing the highest-scoring models, so its identification probability typically increases steadily and reaches $100\%$ with fewer labels than the strongest baseline. In contrast, \textsc{Random}, \textsc{VMA}, and \textsc{AMC} generally improve more slowly, while \textsc{Margin} and \textsc{Min Agreement} often exhibit dataset-specific oscillations.

\subsubsection{Annotation Efficiency for Near-Best Models}

Table~\ref{tab:near_best_label_efficiency} shows that the gains of \framework\ persist when the goal is relaxed from identifying the best model to identifying a near-best model. Across the $69$ dataset-threshold combinations, \framework\ uses fewer annotations in $68$ settings, with the remaining setting using almost the same number of annotations as the strongest baseline. The largest reduction is $84.78\%$ on XSum (ROUGE-L) at $\delta=0.1\%$.

These results indicate that the queries selected by \framework\ support not only accurate best model identification, but also narrow the candidate set quickly enough to recover a near-best model with very few annotations. The same qualitative pattern across all three values of $\delta$ indicates that the advantage of \framework\ is not tied to a single near-best threshold.

\begin{table}[!ht]
    \centering
    \resizebox{0.6\linewidth}{!}{
\begin{tabular}{lccc}
\specialrule{1.5pt}{0pt}{0pt}
Dataset & $\delta = 0.1\%$ & $\delta = 0.5\%$ & $\delta = 1\%$ \\
\midrule
bAbI & $\downarrow\mathbf{50.74}\%$ & $\downarrow\mathbf{47.64}\%$ & $\downarrow\mathbf{50.25}\%$ \\
BBH & $\downarrow\mathbf{28.71}\%$ & $\downarrow\mathbf{39.19}\%$ & $\downarrow\mathbf{35.10}\%$ \\
CNN/DailyMail (ROUGE-L) & $\downarrow\mathbf{69.15}\%$ & $\downarrow\mathbf{69.15}\%$ & $\downarrow\mathbf{69.15}\%$ \\
CNN/DailyMail (BERTScore) & $\downarrow\mathbf{78.99}\%$ & $\downarrow\mathbf{77.14}\%$ & $\downarrow\mathbf{72.73}\%$ \\
Flickr30k & $\downarrow\mathbf{38.05}\%$ & $\downarrow\mathbf{38.05}\%$ & $\downarrow\mathbf{38.05}\%$ \\
GPQA & $\downarrow\mathbf{43.78}\%$ & $\downarrow\mathbf{43.66}\%$ & $\downarrow\mathbf{43.43}\%$ \\
GSM8K & $\downarrow\mathbf{63.31}\%$ & $\downarrow\mathbf{63.17}\%$ & $\downarrow\mathbf{62.58}\%$ \\
IMDb & $\downarrow\mathbf{61.74}\%$ & $\downarrow\mathbf{60.44}\%$ & $\downarrow\mathbf{58.62}\%$ \\
MATH (Chain-of-Thought) & $\downarrow\mathbf{67.20}\%$ & $\downarrow\mathbf{67.20}\%$ & $\downarrow\mathbf{61.04}\%$ \\
MATH & $\downarrow\mathbf{31.72}\%$ & $\downarrow\mathbf{31.72}\%$ & $\downarrow\mathbf{31.72}\%$ \\
MMLU-Pro & $\downarrow\mathbf{29.87}\%$ & $\downarrow\mathbf{39.64}\%$ & $\downarrow\mathbf{27.23}\%$ \\
MuSR & $\downarrow\mathbf{31.40}\%$ & $\downarrow\mathbf{31.40}\%$ & $\downarrow\mathbf{31.40}\%$ \\
NarrativeQA & $\downarrow\mathbf{58.43}\%$ & $\downarrow\mathbf{56.14}\%$ & $\downarrow\mathbf{58.56}\%$ \\
Natural Questions (Closed-Book) & $\downarrow\mathbf{29.78}\%$ & $\downarrow\mathbf{31.94}\%$ & $\downarrow\mathbf{24.48}\%$ \\
Natural Questions (Open-Book) & $\downarrow\mathbf{57.53}\%$ & $\downarrow\mathbf{50.28}\%$ & $\uparrow0.45\%$ \\
VQAv2 & $\downarrow\mathbf{38.32}\%$ & $\downarrow\mathbf{34.41}\%$ & $\downarrow\mathbf{31.51}\%$ \\
WMT14 (Czech-English) & $\downarrow\mathbf{47.74}\%$ & $\downarrow\mathbf{32.35}\%$ & $\downarrow\mathbf{30.85}\%$ \\
WMT14 (German-English) & $\downarrow\mathbf{17.10}\%$ & $\downarrow\mathbf{25.73}\%$ & $\downarrow\mathbf{35.23}\%$ \\
WMT14 (French-English) & $\downarrow\mathbf{44.68}\%$ & $\downarrow\mathbf{47.41}\%$ & $\downarrow\mathbf{40.26}\%$ \\
WMT14 (Hindi-English) & $\downarrow\mathbf{46.97}\%$ & $\downarrow\mathbf{43.67}\%$ & $\downarrow\mathbf{9.49}\%$ \\
WMT14 (Russian-English) & $\downarrow\mathbf{15.61}\%$ & $\downarrow\mathbf{19.96}\%$ & $\downarrow\mathbf{21.16}\%$ \\
XSum (ROUGE-L) & $\downarrow\mathbf{84.78}\%$ & $\downarrow\mathbf{84.73}\%$ & $\downarrow\mathbf{84.66}\%$ \\
XSum (BERTScore) & $\downarrow\mathbf{50.00}\%$ & $\downarrow\mathbf{50.00}\%$ & $\downarrow\mathbf{50.00}\%$ \\
\specialrule{1.5pt}{0pt}{0pt}
\end{tabular}
    }
    \caption{Annotation efficiency for near-best model selection across datasets. Each $\delta$ denotes a relative gap from the best model. Bolded $\downarrow$ values indicate fewer labels than the strongest baseline; $\uparrow$ indicates more labels.}
    
  \vspace{-5pt}
    \label{tab:near_best_label_efficiency}
\end{table}

\subsubsection{Robustness Analysis}

Table~\ref{tab:gap_percentile} further examines robustness by reporting the $95$th percentile performance gap at the budgets where \framework\ reaches $70\%$, $80\%$, $90\%$, and $100\%$ identification probability. Each method is evaluated at the same budget in a given row, so the comparison isolates how close the selected model remains to the true best model under the same annotation cost. Under this criterion, \framework\ has the lowest gap in $64$ of the $92$ comparisons across datasets and identification levels, and ranks among the top two methods in $79$ of $92$ comparisons.

The low gaps for \framework\ mean that large performance losses remain rare, even in the more difficult realizations summarized by this metric. When the selected model differs from the true best, its average score is still close to the best score, so selection errors have limited impact on downstream model quality. This table shows that \framework\ remains robust even in worst case scenarios.

\begin{table*}[!ht]
    \centering
    \resizebox{1\textwidth}{!}{
\begin{tabular}{l c c c c c c}
\specialrule{1.5pt}{0pt}{0pt}
Dataset & \framework\ & \textsc{Random} & \textsc{Margin} & \textsc{Min Agreement} & \textsc{VMA} & \textsc{AMC} \\
Ident. prob. & \small(70\%/80\%/90\%/100\%) & \small(70\%/80\%/90\%/100\%) & \small(70\%/80\%/90\%/100\%) & \small(70\%/80\%/90\%/100\%) & \small(70\%/80\%/90\%/100\%) & \small(70\%/80\%/90\%/100\%) \\
\midrule
bAbI & $\mathbf{2.00}$/$\mathbf{1.40}$/$\mathbf{0.60}$/$\mathbf{0.00}$ & $2.60$/$2.60$/$2.60$/$2.20$ & $2.60$/$2.60$/$2.60$/$2.20$ & $2.60$/$2.60$/$2.60$/$2.20$ & $\underline{2.40}$/$\underline{2.40}$/$\underline{2.20}$/$\underline{2.00}$ & $2.60$/$\underline{2.40}$/$2.40$/$2.20$ \\
BBH & $3.00$/$2.10$/$\mathbf{0.80}$/$\mathbf{0.00}$ & $2.90$/$2.60$/$2.30$/$0.90$ & $2.80$/$2.60$/$2.40$/$0.80$ & $\mathbf{2.10}$/$\underline{2.00}$/$1.80$/$\underline{0.10}$ & $\mathbf{2.10}$/$\mathbf{1.90}$/$\underline{1.60}$/$0.20$ & $\underline{2.20}$/$\underline{2.00}$/$\underline{1.60}$/$0.30$ \\
CNN/DailyMail (ROUGE-L) & $\mathbf{1.25}$/$\mathbf{1.25}$/$\mathbf{1.05}$/$\mathbf{0.00}$ & $7.50$/$7.50$/$2.92$/$1.91$ & $7.62$/$7.62$/$3.44$/$3.02$ & $\underline{3.21}$/$\underline{3.21}$/$\underline{2.19}$/$\underline{1.34}$ & $7.47$/$7.47$/$3.13$/$1.95$ & $7.57$/$7.57$/$3.01$/$1.83$ \\
CNN/DailyMail (BERTScore) & $\mathbf{2.63}$/$\mathbf{1.64}$/$\mathbf{1.18}$/$\mathbf{0.00}$ & $7.55$/$7.37$/$7.18$/$\mathbf{0.00}$ & $5.75$/$4.49$/$2.63$/$\mathbf{0.00}$ & $\underline{3.24}$/$\underline{3.28}$/$\underline{2.27}$/$\mathbf{0.00}$ & $7.39$/$6.89$/$6.27$/$\mathbf{0.00}$ & $7.44$/$7.34$/$6.97$/$\mathbf{0.00}$ \\
Flickr30k & $\mathbf{1.63}$/$\mathbf{1.37}$/$\mathbf{1.13}$/$\mathbf{0.00}$ & $2.54$/$2.12$/$\underline{1.80}$/$\mathbf{0.00}$ & $3.16$/$2.51$/$2.33$/$\mathbf{0.00}$ & $\underline{1.89}$/$\underline{1.70}$/$2.13$/$\mathbf{0.00}$ & $2.59$/$2.05$/$1.91$/$\mathbf{0.00}$ & $2.53$/$2.06$/$1.91$/$\mathbf{0.00}$ \\
GPQA & $10.40$/$\mathbf{1.80}$/$\mathbf{0.60}$/$\mathbf{0.00}$ & $9.80$/$\underline{2.60}$/$2.20$/$1.20$ & $\underline{9.60}$/$\underline{2.60}$/$2.00$/$\underline{1.00}$ & $10.20$/$3.20$/$2.40$/$1.40$ & $\mathbf{9.00}$/$2.80$/$\underline{1.80}$/$\underline{1.00}$ & $\underline{9.60}$/$2.80$/$\underline{1.80}$/$\underline{1.00}$ \\
GSM8K & $14.40$/$\underline{2.20}$/$\mathbf{0.60}$/$\mathbf{0.00}$ & $\underline{2.40}$/$2.40$/$2.20$/$2.00$ & $\underline{2.40}$/$\underline{2.20}$/$2.20$/$2.20$ & $\underline{2.40}$/$\underline{2.20}$/$\underline{2.00}$/$\underline{1.40}$ & $\mathbf{2.00}$/$\mathbf{2.00}$/$\underline{2.00}$/$1.80$ & $2.60$/$2.40$/$2.40$/$2.20$ \\
IMDb & $\mathbf{1.60}$/$\mathbf{0.80}$/$\mathbf{0.40}$/$\mathbf{0.00}$ & $6.20$/$6.00$/$5.40$/$\underline{0.60}$ & $5.80$/$5.60$/$5.60$/$\underline{0.60}$ & $\underline{2.60}$/$\underline{2.40}$/$\underline{2.20}$/$\mathbf{0.00}$ & $3.20$/$3.00$/$2.80$/$\mathbf{0.00}$ & $4.20$/$3.60$/$3.40$/$\mathbf{0.00}$ \\
MATH (Chain-of-Thought) & $\mathbf{2.50}$/$\mathbf{2.00}$/$\mathbf{1.00}$/$\mathbf{0.00}$ & $3.50$/$3.50$/$3.50$/$3.00$ & $4.50$/$4.00$/$4.00$/$4.00$ & $\mathbf{2.50}$/$\underline{2.50}$/$\underline{2.50}$/$\underline{2.50}$ & $3.50$/$3.50$/$3.50$/$3.00$ & $\underline{3.00}$/$3.50$/$3.00$/$3.00$ \\
MATH & $\mathbf{4.50}$/$\mathbf{3.50}$/$2.50$/$\mathbf{0.00}$ & $8.17$/$7.00$/$2.50$/$1.50$ & $\underline{7.00}$/$5.50$/$\underline{2.00}$/$1.00$ & $\mathbf{4.50}$/$\underline{4.50}$/$\mathbf{1.50}$/$\underline{0.50}$ & $7.17$/$6.67$/$2.50$/$1.50$ & $\underline{7.00}$/$5.67$/$3.00$/$1.50$ \\
MMLU-Pro & $9.80$/$\underline{1.70}$/$\underline{0.80}$/$\mathbf{0.00}$ & $\underline{3.80}$/$2.00$/$1.70$/$\mathbf{0.00}$ & $\underline{3.80}$/$1.90$/$1.60$/$\mathbf{0.00}$ & $4.00$/$\mathbf{0.50}$/$\mathbf{0.30}$/$\mathbf{0.00}$ & $\mathbf{3.60}$/$2.00$/$1.60$/$\mathbf{0.00}$ & $3.90$/$1.90$/$1.50$/$\mathbf{0.00}$ \\
MuSR & $\underline{11.00}$/$\underline{9.67}$/$\underline{6.67}$/$\mathbf{0.00}$ & $16.67$/$14.33$/$12.67$/$\mathbf{0.00}$ & $16.00$/$14.67$/$12.33$/$\mathbf{0.00}$ & $\mathbf{8.33}$/$\mathbf{5.33}$/$\mathbf{5.67}$/$\mathbf{0.00}$ & $16.00$/$14.33$/$12.00$/$\mathbf{0.00}$ & $17.00$/$14.67$/$13.33$/$\mathbf{0.00}$ \\
NarrativeQA & $4.40$/$\mathbf{2.14}$/$\mathbf{1.10}$/$\mathbf{0.00}$ & $4.32$/$2.89$/$2.72$/$0.66$ & $5.81$/$4.24$/$3.75$/$\mathbf{0.00}$ & $\mathbf{2.59}$/$\underline{2.26}$/$\underline{2.32}$/$2.10$ & $4.56$/$2.89$/$2.45$/$\underline{0.35}$ & $\underline{3.53}$/$2.83$/$2.37$/$0.90$ \\
Natural Questions (Closed-Book) & $\underline{4.30}$/$\underline{2.69}$/$\underline{1.47}$/$\mathbf{0.00}$ & $4.85$/$3.75$/$2.29$/$\mathbf{0.00}$ & $4.72$/$3.29$/$2.17$/$\mathbf{0.00}$ & $\mathbf{3.30}$/$\mathbf{1.67}$/$\mathbf{0.92}$/$\mathbf{0.00}$ & $5.21$/$4.14$/$2.47$/$\mathbf{0.00}$ & $5.10$/$3.41$/$2.15$/$\mathbf{0.00}$ \\
Natural Questions (Open-Book) & $\underline{5.44}$/$\underline{4.92}$/$\mathbf{3.59}$/$\mathbf{0.00}$ & $9.13$/$8.70$/$8.17$/$1.07$ & $8.90$/$8.43$/$7.49$/$1.28$ & $\mathbf{4.95}$/$\mathbf{4.78}$/$\underline{4.61}$/$\mathbf{0.00}$ & $8.67$/$8.09$/$7.53$/$1.64$ & $8.63$/$8.02$/$6.96$/$\underline{0.79}$ \\
VQAv2 & $\mathbf{9.20}$/$\mathbf{3.40}$/$\mathbf{0.60}$/$\mathbf{0.00}$ & $11.60$/$6.80$/$3.00$/$2.00$ & $11.40$/$6.20$/$2.80$/$2.20$ & $\underline{10.40}$/$\underline{3.80}$/$\underline{2.60}$/$\underline{1.40}$ & $11.40$/$6.60$/$3.00$/$2.40$ & $11.20$/$6.00$/$2.80$/$2.00$ \\
WMT14 (Czech-English) & $\underline{2.08}$/$\underline{1.92}$/$\mathbf{0.38}$/$\mathbf{0.00}$ & $3.43$/$3.24$/$2.27$/$\underline{0.22}$ & $4.19$/$3.93$/$3.30$/$0.43$ & $\mathbf{2.07}$/$\mathbf{1.84}$/$\underline{1.28}$/$\mathbf{0.00}$ & $3.61$/$3.39$/$2.50$/$0.27$ & $3.70$/$3.36$/$2.29$/$\underline{0.22}$ \\
WMT14 (German-English) & $\mathbf{2.07}$/$\mathbf{1.44}$/$\mathbf{0.15}$/$\mathbf{0.00}$ & $3.23$/$2.15$/$0.83$/$0.13$ & $3.49$/$2.56$/$1.44$/$0.06$ & $\underline{2.62}$/$\underline{1.50}$/$\underline{0.17}$/$\underline{0.05}$ & $3.06$/$2.10$/$0.88$/$0.08$ & $2.90$/$2.18$/$0.78$/$0.07$ \\
WMT14 (French-English) & $2.41$/$\mathbf{0.21}$/$\mathbf{0.08}$/$\mathbf{0.00}$ & $\underline{1.60}$/$1.34$/$0.96$/$0.27$ & $1.92$/$1.59$/$1.11$/$0.36$ & $\mathbf{0.44}$/$\underline{0.38}$/$\underline{0.23}$/$\underline{0.19}$ & $1.69$/$1.40$/$0.96$/$0.24$ & $1.75$/$1.42$/$0.93$/$0.26$ \\
WMT14 (Hindi-English) & $1.68$/$1.27$/$\mathbf{0.26}$/$\mathbf{0.00}$ & $1.75$/$1.70$/$0.82$/$0.27$ & $\underline{1.22}$/$\mathbf{0.92}$/$\underline{0.56}$/$0.26$ & $\mathbf{0.59}$/$\underline{0.92}$/$1.25$/$\underline{0.10}$ & $1.76$/$1.52$/$1.06$/$0.28$ & $1.57$/$1.51$/$0.66$/$0.27$ \\
WMT14 (Russian-English) & $\underline{1.23}$/$\underline{0.56}$/$\mathbf{0.13}$/$\mathbf{0.00}$ & $1.73$/$1.30$/$0.92$/$0.12$ & $1.88$/$1.66$/$1.46$/$\underline{0.04}$ & $\mathbf{0.29}$/$\mathbf{0.27}$/$\underline{0.26}$/$0.06$ & $1.75$/$1.39$/$1.07$/$0.11$ & $1.72$/$1.40$/$0.88$/$0.08$ \\
XSum (ROUGE-L) & $7.66$/$4.59$/$3.94$/$\mathbf{0.00}$ & $\underline{4.92}$/$4.20$/$\underline{1.96}$/$1.36$ & $\mathbf{2.12}$/$\mathbf{1.99}$/$\mathbf{1.92}$/$1.68$ & $8.92$/$\underline{3.74}$/$2.05$/$\underline{1.32}$ & $4.97$/$4.19$/$2.07$/$1.43$ & $5.08$/$4.24$/$1.98$/$1.35$ \\
XSum (BERTScore) & $\mathbf{3.69}$/$\mathbf{3.50}$/$\mathbf{3.15}$/$\mathbf{0.00}$ & $3.94$/$3.77$/$3.52$/$\underline{1.83}$ & $3.82$/$3.77$/$3.56$/$\mathbf{0.00}$ & $4.06$/$3.87$/$3.41$/$\mathbf{0.00}$ & $3.88$/$3.74$/$3.49$/$2.06$ & $\underline{3.76}$/$\underline{3.62}$/$\underline{3.34}$/$1.93$ \\
\specialrule{1.5pt}{0pt}{0pt}
\end{tabular}
    }
    \caption{95th percentile performance gap (\%) at the budget where \framework\ reaches identification probabilities 70\%/80\%/90\%/100\% on each dataset. Best values are bold; second-best are underlined.}
      \vspace{-5pt}
    \label{tab:gap_percentile}
\end{table*}

\section{Discussion}
\label{sec:discussion}
Our results show that \framework\ works well across a broad range of LLM tasks, datasets, and evaluation metrics. Across $23$ datasets covering reasoning, question answering, language understanding, summarization, image captioning, math, classification, visual question answering, and machine translation, \framework\ consistently reduces annotation cost for best model identification and improves near-best model selection. These gains persist across diverse reference-based metrics, including BERTScore, ROUGE, BLEU, F1, mathematical equivalence, and exact match, indicating that the method is not tied to a particular task family or scoring rule.

More broadly, \framework\ provides a practical approach to active model selection in realistic black-box settings with limited annotation budgets. Because it relies only on model outputs and an external oracle, it applies to both open-source and API-only models.

\paragraph{Limitations and future work.}
Our study focuses on reference-based evaluation with a reliable oracle signal, and does not cover direct human preferences, LLM-as-a-Judge systems, or other noisy and potentially biased feedback sources. We also assume that the candidate LLMs and query pool are fixed before selection begins. Future work could extend \framework\ to these evaluation protocols, dynamic model pools or streaming query distributions, adaptive stopping rules, and theoretical guarantees for the selection rule.

\paragraph{Broader impacts.}
By reducing annotation cost, \framework\ can make rigorous comparison of large language models more accessible. However, biased annotations, candidate models, or similarity measures may lead to biased model choices, and limited evaluation may select models that fail on unmeasured behaviors, especially in safety-critical domains. More efficient selection could also be misused to optimize models for harmful applications such as persuasive generation, spam, or disinformation. These risks motivate careful choice of annotation protocols, evaluation criteria, and deployment contexts.

\clearpage
\bibliographystyle{plain}
\bibliography{references}

\newpage
\appendix

\section{Derivation of the Query Selection Rule}
\label{sec:appendix-derivation}

We provide the full derivation of the tractable query selection rule used in Section~\ref{sec:llm-selector}. Starting from the mutual information objective in~\Eqref{eq: mi-expanded}, the derivation replaces the intractable summation over the reference space with a posterior-weighted pairwise similarity objective over candidate model responses. The auxiliary algebraic and geometric steps used in the derivation are collected in Section~\ref{sec:aux-derivations}, the assumptions behind the approximations are discussed in Section~\ref{subsec:method-approximation-analysis}, and the resulting selection rule is evaluated with synthetic data in Section~\ref{subsec:synthetic-final-rule-validation}.

We start from the mutual information objective derived in~\Eqref{eq: mi-expanded}:
\begin{equation*}
\mathbb{I}(F;R \mid A_t,q)
= \sum_{j=1}^m \sum_{r \in \mathcal{R}}
\mathbb{P}(F=f_j \mid A_t)\mathbb{P}(R=r \mid F=f_j,q)
\log
\frac{\mathbb{P}(R=r \mid F=f_j,q)}
{\mathbb{P}(R=r \mid A_t,q)}.
\end{equation*}

As the first step, we apply a second-order Taylor approximation of $x\log x$ around $x=1$ to the logarithmic term. Applying this approximation, we obtain

\begin{align*}
\mathbb{I}(F;R \mid A_t,q)
&\approx
\sum_{j=1}^m \mathbb{P}(F = f_j \mid A_t)
\sum_{r \in \mathcal{R}} \mathbb{P}(R = r \mid A_t, q)
\Bigg[
\left(
\frac{\mathbb{P}(R = r \mid F = f_j, q)}{\mathbb{P}(R = r \mid A_t, q)} - 1
\right) \\
&\hspace{4.0cm}
+ \frac{1}{2}
\left(
\frac{\mathbb{P}(R = r \mid F = f_j, q)}{\mathbb{P}(R = r \mid A_t, q)} - 1
\right)^2
\Bigg].
\end{align*}

The linear term vanishes for every $j$, since
 $\sum_{r \in \mathcal{R}} \mathbb{P}(R = r \mid A_t, q)
\left(
\frac{\mathbb{P}(R = r \mid F = f_j, q)}{\mathbb{P}(R = r \mid A_t, q)} - 1\right) = 0.$
Therefore,
\begin{equation*}
\mathbb{I}(F;R \mid A_t,q)
\approx
\frac{1}{2}
\sum_{j=1}^m \mathbb{P}(F = f_j \mid A_t)
\sum_{r \in \mathcal{R}} \mathbb{P}(R = r \mid A_t, q)
\left(
\frac{\mathbb{P}(R = r \mid F = f_j, q)}{\mathbb{P}(R = r \mid A_t, q)} - 1
\right)^2.
\end{equation*}

Next, using the weighted variance identity, we rewrite the above expression in pairwise form:
\begin{align}
\mathbb{I}(F;R \mid A_t,q)
&\approx
\frac{1}{4}
\sum_{j=1}^m \sum_{k=1}^m
\mathbb{P}(F = f_j \mid A_t)\mathbb{P}(F = f_k \mid A_t)
\nonumber \\
&\quad \cdot
\sum_{r \in \mathcal{R}} \mathbb{P}(R = r \mid A_t, q)
\left(
\frac{\mathbb{P}(R = r \mid F = f_j, q)}{\mathbb{P}(R = r \mid A_t, q)}
-
\frac{\mathbb{P}(R = r \mid F = f_k, q)}{\mathbb{P}(R = r \mid A_t, q)}
\right)^2.
\label{eq: pairwise-discrete}
\end{align}

We now connect this expression to the similarity score $s$. Given an observed annotation $r$ for query $q$, the posterior over models is updated as
\begin{equation*}
\mathbb{P}(F = f_j \mid A_t \cup \{(q,r)\})
\propto
\mathbb{P}(F = f_j \mid A_t)\exp\!\left(\frac{s(r,f_j(q))}{\tau}\right).
\end{equation*}

Applying a first-order Taylor approximation to the exponential term and linearizing yields
\begin{equation*}
\mathbb{P}(R = r \mid F = f_j, q)
-
\mathbb{P}(R = r \mid F = f_k, q)
\approx
\frac{1}{Z(\tau)\tau}
\big(s(r,f_j(q)) - s(r,f_k(q))\big).
\end{equation*}

Substituting this into~\Eqref{eq: pairwise-discrete} and dropping multiplicative constants that do not affect the optimizer yields
\begin{align}
q_t
=
\argmax_{q \in U_t}
\sum_{j=1}^m \sum_{k=1}^m
\mathbb{P}(F = f_j \mid A_t)\mathbb{P}(&F = f_k \mid A_t)
\sum_{r \in \mathcal{R}} \mathbb{P}(R = r \mid A_t, q)\nonumber\\
&\cdot\big(s(r,f_j(q)) - s(r,f_k(q))\big)^2.
\label{eq: score-difference-discrete}
\end{align}

To simplify this expression further, we assume that the similarity function admits a feature representation of the form
$s(x,y) = \langle \Phi(x), \Phi(y) \rangle$,
where $\Phi : \mathcal{R} \to \mathbb{R}^d$ is a feature map satisfying $\|\Phi(x)\|_2 = 1$ for all $x \in \mathcal{R}$.
Under these assumptions,
\begin{equation}
\sum_{r \in \mathcal{R}} \mathbb{P}(R = r \mid A_t, q)
\big(s(r,f_j(q)) - s(r,f_k(q))\big)^2
\approx
2c\big(1 - s(f_j(q), f_k(q))\big),
\label{eq: geometric-simplification-discrete}
\end{equation}
for some constant $c > 0$. Substituting~\Eqref{eq: geometric-simplification-discrete} into~\Eqref{eq: score-difference-discrete}, we obtain
\begin{equation*}
q_t
=
\argmax_{q \in U_t}
\sum_{j=1}^m \sum_{k=1}^m
\mathbb{P}(F = f_j \mid A_t)\mathbb{P}(F = f_k \mid A_t)
\, 2c \big(1 - s(f_j(q), f_k(q))\big).
\end{equation*}

Since $2c$ is a positive constant and $\sum_{j=1}^m \sum_{k=1}^m
\mathbb{P}(F = f_j \mid A_t)\mathbb{P}(F = f_k \mid A_t) = 1,$
the additive and multiplicative constants can be dropped.  
Thus, the selection rule reduces to~\Eqref{eq: final-query-selection-rule}:

\[
q_t
=
\argmin_{q \in U_t}
\sum_{j=1}^m \sum_{k=1}^m
\mathbb{P}(F = f_j \mid A_t)\mathbb{P}(F = f_k \mid A_t)
s(f_j(q), f_k(q))
\]

\subsection{Auxiliary Derivations}
\label{sec:aux-derivations}

\paragraph{Second-order Taylor approximation of $x \log x$.}
Let $g(x) = x \log x$. Then
\[
g(1) = 0, \qquad g'(1) = 1, \qquad g''(1) = 1.
\]
Hence, the second-order Taylor expansion around $x=1$ is
\[
x \log x = (x-1) + \frac{1}{2}(x-1)^2 + O((x-1)^3).
\]

\paragraph{Weighted variance identity.}
For any numbers $a_1,\dots,a_m$ and weights $w_1,\dots,w_m$ satisfying $\sum_{j=1}^m w_j = 1$, we have
\[
\sum_{j=1}^m w_j (a_j - \bar a)^2
=
\frac{1}{2}
\sum_{j=1}^m \sum_{k=1}^m
w_j w_k (a_j - a_k)^2,
\qquad
\bar a = \sum_{j=1}^m w_j a_j.
\]
In our case,
\[
a_j =
\frac{\mathbb{P}(R = r \mid F = f_j, q)}{\mathbb{P}(R = r \mid A_t, q)},
\qquad
w_j = \mathbb{P}(F = f_j \mid A_t),
\]
and therefore $\bar a = 1$.

\paragraph{First-order approximation of the likelihood model.}
Applying a first-order Taylor approximation to the exponential term around a common local baseline gives
\[
\exp\!\left(\frac{s(r,f_j(q))}{\tau}\right)
-
\exp\!\left(\frac{s(r,f_k(q))}{\tau}\right)
\approx
\frac{1}{\tau}\big(s(r,f_j(q)) - s(r,f_k(q))\big).
\]
Since the normalization constant is shared, this implies
\[
\mathbb{P}(R = r \mid F = f_j, q)
-
\mathbb{P}(R = r \mid F = f_k, q)
\approx
\frac{1}{Z(\tau)\tau}\big(s(r,f_j(q)) - s(r,f_k(q))\big).
\]

\paragraph{Geometric simplification under the kernel similarity assumption.}
Assume that the similarity function admits a kernel representation of the form
\[
s(x,y) = \langle \Phi(x), \Phi(y) \rangle,
\quad \text{with } \|\Phi(x)\|_2 = 1.
\]
Then
\begin{align*}
&\sum_{r \in \mathcal{R}} \mathbb{P}(R = r \mid A_t, q)
\big(s(r,f_j(q)) - s(r,f_k(q))\big)^2 \\
&\qquad=
\sum_{r \in \mathcal{R}} \mathbb{P}(R = r \mid A_t, q)
\big\langle \Phi(r), \Phi(f_j(q)) - \Phi(f_k(q)) \big\rangle^2.
\end{align*}
Let $v = \Phi(f_j(q)) - \Phi(f_k(q))$. The above expression can be written as
\[
\sum_{r \in \mathcal{R}} \mathbb{P}(R = r \mid A_t, q)
\langle \Phi(r), v \rangle^2
= v^\top \Sigma_q v,
\]
where
\[
\Sigma_q
= \sum_{r \in \mathcal{R}} \mathbb{P}(R = r \mid A_t, q)\,\Phi(r)\Phi(r)^\top.
\]
Since computing $\Sigma_q$ is intractable due to the size of $\mathcal{R}$, we approximate it by a scalar multiple of the identity, yielding
\[
v^\top \Sigma_q v \approx c\,\|v\|_2^2.
\]
Using $\|\Phi(x)\|_2 = 1$, we obtain
\[
\|\Phi(f_j(q)) - \Phi(f_k(q))\|_2^2
= 2\big(1 - s(f_j(q), f_k(q))\big).
\]
This yields the reduction in~\Eqref{eq: geometric-simplification-discrete}.

\subsection{Approximation Analysis}
\label{subsec:method-approximation-analysis}

In this section, we discuss the assumptions and approximations used to derive the tractable query selection rule in Section~\ref{subsec:algorithm}. While these approximations simplify the original information-theoretic objective, they preserve its key structure and enable an efficient implementation.

\paragraph{Second-order approximation of the mutual information objective.}
The mutual information objective involves a logarithmic term that prevents a direct simplification into a tractable form. We approximate this term using a second-order Taylor expansion, which yields a quadratic form and enables a pairwise decomposition in terms of differences between model predictions. This approximation is well-behaved when the model-specific predictive distributions are close to the posterior predictive distribution, which is typically the case when candidate models produce similar responses. As a result, it provides a tractable surrogate that remains sensitive to informative queries.

\paragraph{First-order approximation of the likelihood model.}
The likelihood function is defined via a temperature-scaled exponential of the similarity score. We approximate this exponential using a first-order expansion, which linearizes the dependence on the similarity function. This step allows us to replace probability differences with score differences, making the objective directly interpretable in terms of the similarity metric. Since the exponential function is smooth, this approximation is well-behaved in regimes where score differences are moderate, which is typical when comparing outputs of reasonably strong models.

\paragraph{Kernel representation of the similarity metric.}
We assume that the similarity function admits a kernel representation of the form $s(x,y) = \langle \Phi(x), \Phi(y) \rangle$. This assumption is natural for commonly used text similarity metrics, many of which can be interpreted as inner products in an embedding space. For example, in n-gram matching based metrics, the embedding $\Phi(x)$ can be viewed as a vector of n-gram counts or frequencies, making the similarity equivalent to a dot product between such representations. This representation enables a geometric interpretation of the objective and allows us to express differences in similarity scores in terms of distances between embeddings. Under mild assumptions on the distribution of plausible responses, this further leads to a simple dependence on pairwise similarities between model outputs.

Taken together, these approximations convert the original mutual information objective into a pairwise disagreement measure between models, weighted by their posterior probabilities. This resulting objective retains the core intuition of information maximization: queries are informative when likely models produce diverse responses. At the same time, it avoids the need to explicitly integrate over the large response space, making the method computationally practical.

\subsection{Validation of the Selection Rule with Synthetic Data}
\label{subsec:synthetic-final-rule-validation}

This section evaluates whether the final selection rule preserves the query ordering induced by the mutual information objective in a setting where the objective can be computed exactly. We use the finite set of binary-valued vectors $\mathcal{X}_d=\{0,1\}^d$ as the synthetic response space, and compare vectors with cosine similarity $s(x,y)=x^\top y/(\|x\|_2\|y\|_2)$, setting $s(x,y)=0$ if either vector has zero norm. The goal is deliberately narrow: we do not claim that binary-valued vectors simulate natural language, nor that the approximation is exact for every distribution over the response space. Instead, we ask whether the tractable \textsc{Select-LLM} approximation preserves the query ranking induced by $\mathbb{I}(F;R\mid A_t,q_i)$ when the latter can be computed exactly over a finite support.

This finite set still has a natural connection to text-based evaluation. Consider a simplified language with vocabulary size $d$: vectors in $\mathcal{X}_d$ can be interpreted as feature vectors that record which tokens from this vocabulary are present. This view naturally leads to token matching based text evaluation metrics.

We simulate candidate model sets of size $m\in\{2,5,10,20\}$ with $d=8$, so there are exactly $2^8=256$ possible vectors. We average all results over $10{,}000$ independent seeds. Each seed contains $n_{\mathrm{syn}}=100$ candidate queries. For each query $q_i$ and candidate model $f_j$, we sample the synthetic vector $f_j(q_i)$ uniformly from the nonzero binary-valued vectors. For each seed, we apply~\Eqref{eq: final-query-selection-rule} directly using the synthetic vectors $f_j(q_i)$ and cosine similarity as $s$. We then compare the query ranking induced by this selection rule with the ranking induced by $\mathbb{I}(F;R\mid A_t,q_i)$. We report \textit{top-1 recall}, the fraction of seeds in which the selection rule recovers the query maximizing $\mathbb{I}(F;R\mid A_t,q_i)$; \textit{top-5\% recall}, the fraction of seeds in which the selection rule recovers a query among the top $5\%$ under $\mathbb{I}(F;R\mid A_t,q_i)$; \textit{Spearman correlation}, the rank correlation between the two query rankings; and \textit{pairwise order accuracy}, the fraction of query pairs ordered consistently by the two rankings.

\begin{table}[t]
    \centering
    \small
    \setlength{\tabcolsep}{3.5pt}
    \begin{minipage}{0.49\textwidth}
    \centering
    \begin{tabular}{@{}c|cccc@{}}
    \toprule
    \multicolumn{5}{c}{2 models} \\
    \midrule
    $\max P_j$ & Top-1 & Top-5\% & Spearman & Pairwise \\
    \midrule
    $0.50$ & $1.000$ & $1.000$ & $0.996$ & $0.984$ \\
    $0.60$ & $1.000$ & $1.000$ & $0.995$ & $0.981$ \\
    $0.70$ & $1.000$ & $1.000$ & $0.996$ & $0.982$ \\
    $0.80$ & $1.000$ & $1.000$ & $0.995$ & $0.982$ \\
    $0.90$ & $1.000$ & $1.000$ & $0.995$ & $0.981$ \\
    \bottomrule
    \end{tabular}
    \end{minipage}
    \hfill
    \begin{minipage}{0.49\textwidth}
    \centering
    \begin{tabular}{@{}c|cccc@{}}
    \toprule
    \multicolumn{5}{c}{5 models} \\
    \midrule
    $\max P_j$ & Top-1 & Top-5\% & Spearman & Pairwise \\
    \midrule
    $0.20$ & $0.912$ & $1.000$ & $0.997$ & $0.982$ \\
    $0.35$ & $0.902$ & $1.000$ & $0.997$ & $0.982$ \\
    $0.50$ & $0.889$ & $0.999$ & $0.997$ & $0.982$ \\
    $0.70$ & $0.873$ & $0.999$ & $0.997$ & $0.982$ \\
    $0.90$ & $0.861$ & $0.999$ & $0.997$ & $0.983$ \\
    \bottomrule
    \end{tabular}
    \end{minipage}

    \vspace{0.8em}

    \begin{minipage}{0.49\textwidth}
    \centering
    \begin{tabular}{@{}c|cccc@{}}
    \toprule
    \multicolumn{5}{c}{10 models} \\
    \midrule
    $\max P_j$ & Top-1 & Top-5\% & Spearman & Pairwise \\
    \midrule
    $0.10$ & $0.915$ & $1.000$ & $0.997$ & $0.983$ \\
    $0.30$ & $0.889$ & $1.000$ & $0.997$ & $0.983$ \\
    $0.50$ & $0.869$ & $0.999$ & $0.997$ & $0.983$ \\
    $0.70$ & $0.854$ & $0.999$ & $0.998$ & $0.984$ \\
    $0.90$ & $0.836$ & $0.999$ & $0.998$ & $0.985$ \\
    \bottomrule
    \end{tabular}
    \end{minipage}
    \hfill
    \begin{minipage}{0.49\textwidth}
    \centering
    \begin{tabular}{@{}c|cccc@{}}
    \toprule
    \multicolumn{5}{c}{20 models} \\
    \midrule
    $\max P_j$ & Top-1 & Top-5\% & Spearman & Pairwise \\
    \midrule
    $0.05$ & $0.920$ & $1.000$ & $0.998$ & $0.983$ \\
    $0.30$ & $0.878$ & $1.000$ & $0.998$ & $0.984$ \\
    $0.50$ & $0.864$ & $0.999$ & $0.998$ & $0.986$ \\
    $0.70$ & $0.856$ & $0.999$ & $0.998$ & $0.987$ \\
    $0.90$ & $0.845$ & $0.999$ & $0.998$ & $0.987$ \\
    \bottomrule
    \end{tabular}
    \end{minipage}

    \caption{Synthetic validation in $\{0,1\}^8$ with $n_{\mathrm{syn}}=100$ candidate queries per seed, $\tau=1$, and $10{,}000$ independent seeds. The mutual information $\mathbb{I}(F;R\mid A_t,q_i)$ is computed by summing over all $2^8=256$ binary-valued vectors. We write $P_j=\mathbb{P}(F=f_j\mid A_t)$; the first column reports the largest current model posterior probability. The remaining columns report top-1 recall, top-5\% recall, Spearman rank correlation, and pairwise order accuracy. For $m>2$, the remaining posterior mass is distributed uniformly over the other $m-1$ models.}
    \label{tab:synthetic-final-rule-eig}
    \vspace{-5pt}
\end{table}

\begin{figure}[t]
    \centering
    \includegraphics[width=0.88\textwidth]{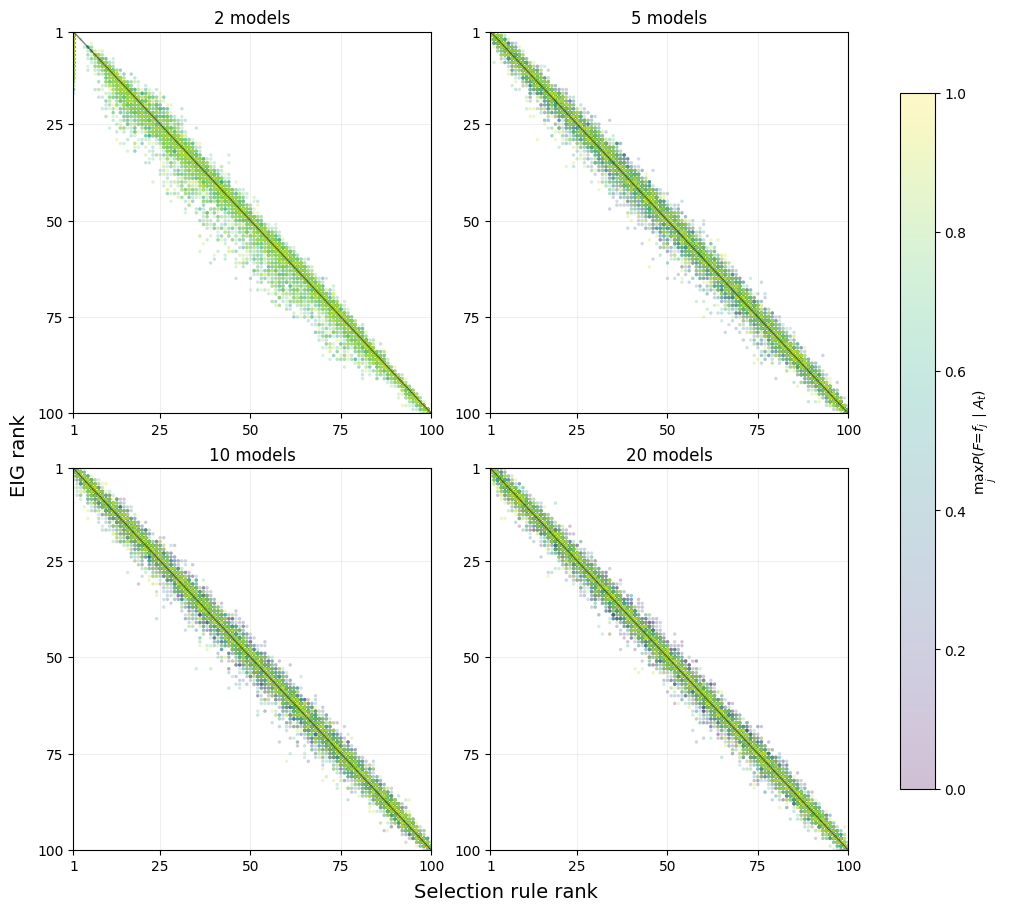}
    \caption{Query-rank comparison between the selection rule and exact mutual information in the synthetic binary-valued vector validation. Each plot corresponds to a different number of candidate models. The $x$-axis gives the rank induced by~\Eqref{eq: final-query-selection-rule}, and the $y$-axis gives the rank induced by $\mathbb{I}(F;R\mid A_t,q_i)$, with rank $1$ at the top. Points near the diagonal indicate agreement between the two rankings.}
    \label{fig:synthetic-final-rule-rank-scatter}
    \vspace{-5pt}
\end{figure}

Table~\ref{tab:synthetic-final-rule-eig} shows that the selection rule strongly preserves the mutual information ranking in this finite binary vector space across different numbers of candidate models. Spearman correlation remains between $0.995$ and $0.998$ in all settings, and pairwise order accuracy remains at least $0.981$. Top-5\% recall is at least $0.999$ in every row, meaning that the selection rule almost always recovers a query in the highest-information part of the exact mutual information ranking. Figure~\ref{fig:synthetic-final-rule-rank-scatter} provides the same evidence: the rank pairs concentrate tightly along the diagonal, indicating that the selection rule and exact mutual information induce highly aligned query orderings.

The top-1 recall results show a mild dependence on the number of candidate models and on the current best model posterior. For $m=2$, top-1 recall is $1.000$ across posterior settings. For larger model sets, top-1 recall remains high but becomes slightly more stringent: it ranges from $0.861$ to $0.912$ for $m=5$, from $0.836$ to $0.915$ for $m=10$, and from $0.845$ to $0.920$ for $m=20$. Within each model count, the strongest top-1 recall usually occurs near the uniform posterior and decreases moderately as the current best model posterior becomes more concentrated. This is also the regime in which the selection process is closer to completion: once one model already has high posterior probability, the best model is more likely to have been identified, and the marginal importance of selecting the single most informative remaining query is reduced. Thus, the selection rule is most accurate in the early, high-uncertainty regime where query choice is most consequential.

\paragraph{Scope of the validation.}
The synthetic validation is a controlled assessment of the acquisition rule, not a proof for arbitrary open-ended text distributions. The results support the narrower claim that, over a finite set where information gain can be computed exactly, the tractable \textsc{Select-LLM} approximation preserves the high-information part of the query ranking induced by mutual information.

\section{Datasets and Model Collections}
\label{sec:appendix-datasets-model-collections}

We evaluate on the following datasets in our experiments: bAbI \cite{weston2015towards} is a suite of synthetic question-answering tasks designed to test prerequisite reasoning skills such as coreference, induction, and path finding. BBH \cite{suzgun2022challenging} collects challenging BIG-Bench tasks that emphasize complex reasoning. GPQA \cite{rein2023gpqa} contains expert-written graduate-level science questions designed to be difficult even when simple search is available. MMLU-Pro \cite{wang2024mmlupro} extends MMLU \cite{hendrycks2021mmlu} with more challenging, reasoning-oriented multiple-choice questions and a larger answer set. MuSR \cite{sprague2023musr} evaluates multistep reasoning over natural-language narratives. CNN/DailyMail \cite{hermann2015teaching} contains news articles paired with human-written highlights, making it a standard benchmark for summarization, while XSum \cite{narayan-etal-2018-dont} contains BBC articles paired with highly abstractive one-sentence summaries. Flickr30k \cite{flickr30k} contains images paired with human-written captions and is used here as an image captioning benchmark. GSM8K \cite{cobbe2021gsm8k} consists of grade-school math word problems with final numerical answers, while MATH \cite{hendrycks2021measuring} contains competition-style math problems that require more advanced mathematical reasoning. IMDb \cite{maas-etal-2011-learning} is a binary sentiment classification benchmark built from movie reviews. NarrativeQA \cite{kocisky-etal-2018-narrativeqa} evaluates question answering over books and movie scripts, while Natural Questions \cite{kwiatkowski2019natural} contains real Google search queries with answers grounded in Wikipedia and is evaluated in both closed-book and open-book settings. VQAv2 \cite{goyal2017making} is a visual question answering benchmark with image-question-answer triples designed to reduce language-only biases. WMT14 \cite{bojar-etal-2014-findings} contains machine translation test sets, and we evaluate translation into English from Czech, German, French, Hindi, and Russian.

Across these datasets, we use reference-based evaluation as the oracle signal rather than LLM-as-a-Judge. This choice avoids relying on judge-based evaluation protocols that can be vulnerable to attacks \cite{zheng2025cheating,jiang2023promptpacker}, exhibit systematic judgment biases \cite{chen-etal-2024-humans}, and show self-inconsistency across judging runs or prompts \cite{haldar-hockenmaier-2025-rating}.

\begin{figure}[t]
  \centering
  \includegraphics[width=\textwidth]{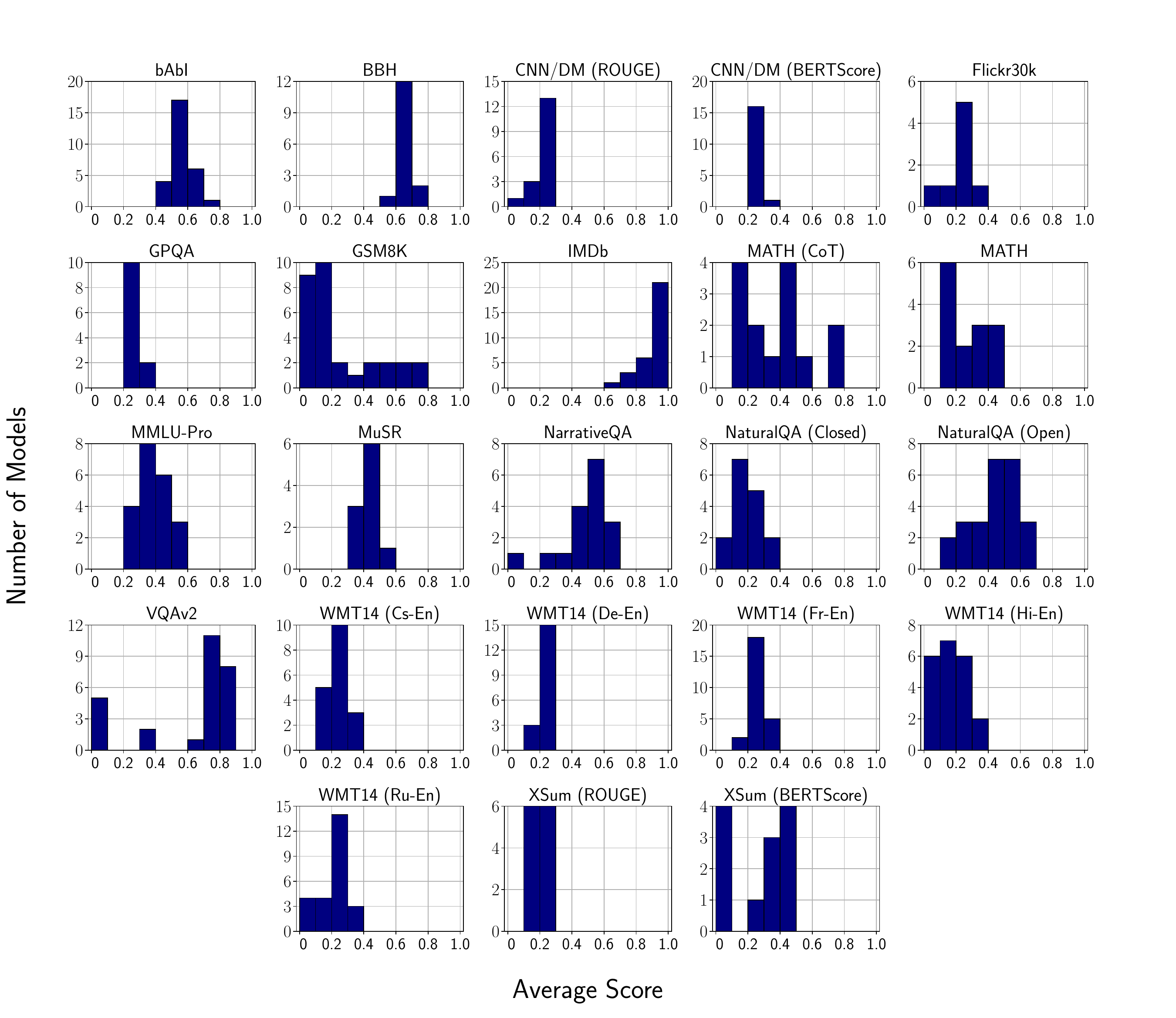}
  \caption{
  Distribution of average model scores for the $23$ datasets used in our experiments.
  Each histogram bins the candidate models by their average oracle score on the corresponding task, using the same task-specific metric as in the main experiments. Scores below zero are included in the first bin.
  }
  \label{fig:dataset_summary_results}
\end{figure}

Figure~\ref{fig:dataset_summary_results} summarizes the performance distribution of the candidate models on each dataset. The histograms show that the model collections vary substantially across tasks: some datasets, such as IMDb, contain many high-performing models, whereas reasoning, question answering, summarization, and translation tasks often have wider or lower-scoring distributions. This variation is important for active model selection because the difficulty of identifying the best model depends not only on the number of candidates, but also on how separated their task-specific performances are.

Table~\ref{tab:dataset_model_collection_summary} provides a complementary summary of the experimental collections, including the selected temperature parameter, dataset sizes, realization sizes, score ranges, model counts, and evaluation metrics.

\begin{table}[t]
    \centering
    \resizebox{\textwidth}{!}{
\begin{tabular}{lcccccc}
\specialrule{1.5pt}{0pt}{0pt}
Dataset & Selected $\tau$ & Dataset size & Realization size $n$ & Avg. score range & Models & Metric \\
\midrule
bAbI & 2.0 & 1,000 & 500 & 0.49--0.71 & 28 & Exact match \\
BBH & 3.0 & 5,761 & 1,000 & 0.59--0.73 & 15 & Exact match \\
CNN/DailyMail (ROUGE-L) & 0.5 & 1,000 & 500 & 0.09--0.27 & 17 & ROUGE-L \\
CNN/DailyMail (BERTScore) & 0.1 & 1,000 & 500 & 0.21--0.30 & 17 & BERTScore \\
Flickr30k & 0.1 & 1,000 & 500 & 0.00--0.31 & 8 & ROUGE-L \\
GPQA & 3.0 & 1,192 & 500 & 0.24--0.39 & 12 & Exact match \\
GSM8K & 1.0 & 1,000 & 500 & 0.06--0.76 & 30 & Exact match \\
IMDb & 0.1 & 1,000 & 500 & 0.65--0.96 & 31 & Exact match \\
MATH (Chain-of-Thought) & 2.0 & 437 & 200 & 0.17--0.78 & 14 & Equivalence \\
MATH & 2.0 & 437 & 200 & 0.16--0.49 & 14 & Equivalence \\
MMLU-Pro & 3.0 & 12,032 & 1,000 & 0.21--0.56 & 21 & Exact match \\
MuSR & 2.0 & 756 & 300 & 0.34--0.54 & 10 & Exact match \\
NarrativeQA & 0.7 & 2,350 & 1,000 & 0.07--0.65 & 17 & F1 \\
Natural Questions (Closed-Book) & 0.1 & 2,876 & 1,000 & 0.04--0.32 & 16 & F1 \\
Natural Questions (Open-Book) & 1.0 & 5,000 & 1,000 & 0.14--0.65 & 25 & F1 \\
VQAv2 & 3.0 & 1,000 & 500 & 0.00--0.86 & 27 & Exact match \\
WMT14 (Czech-English) & 0.6 & 1,000 & 500 & 0.10--0.32 & 18 & BLEU-4 \\
WMT14 (German-English) & 0.9 & 1,000 & 500 & 0.17--0.29 & 18 & BLEU-4 \\
WMT14 (French-English) & 0.1 & 1,000 & 500 & 0.12--0.33 & 25 & BLEU-4 \\
WMT14 (Hindi-English) & 0.3 & 1,000 & 500 & 0.01--0.31 & 21 & BLEU-4 \\
WMT14 (Russian-English) & 0.5 & 1,000 & 500 & 0.02--0.31 & 25 & BLEU-4 \\
XSum (ROUGE-L) & 0.1 & 1,000 & 500 & 0.11--0.30 & 12 & ROUGE-L \\
XSum (BERTScore) & 4.0 & 1,000 & 500 & $-0.28$--0.46 & 12 & BERTScore \\
\specialrule{1.5pt}{0pt}{0pt}
\end{tabular}
    }
    \caption{Summary of the datasets and model collections used in the experiments. Dataset size is the number of instances available in the oracle pool, and realization size $n$ is the number of instances sampled for each experimental realization. Score ranges report the minimum and maximum average model scores under the dataset's evaluation metric.}
    \label{tab:dataset_model_collection_summary}
\end{table}

Across all datasets, our experiments include $156$ unique evaluated models. The model pool includes OpenAI GPT, text, and code models \cite{Brown2020gpt3,Ouyang2022instructgpt,openai2023gpt4,openai2024gpt4o,openaiapi}; Anthropic Claude models \cite{claude2,claude3,anthropicapi}; Google Gemini, PaLM, and PaliGemma models \cite{geminiteam2023gemini,geminiapi,anil2023palm2,beyer2024paligemma}; AI21 Jurassic models \cite{ai21jurassic}; Cohere Command and Command-R models \cite{cohereapi,command-r,command-r-plus}; Aleph Alpha Luminous models \cite{alephalpha_luminous}; Writer Palmyra models \cite{writer_palmyra}; Microsoft TNLGv2 and Phi models \cite{microsoft_tnlg,abdin2024phi3}; Meta LLaMA and Llama 3 models \cite{touvron2023llama,touvron2023llama2,llama3}; Mistral, Mixtral, and Pixtral models \cite{jiang2023mistral,mixtral,pixtral2024}; Qwen, Qwen1.5/Qwen2.5, and Qwen-VL models \cite{bai2023qwen,qwen1.5,bai2023qwenvl}; Gemma models \cite{gemma}; Falcon models \cite{almazrouei2023falcon}; Yi models \cite{young2024yi}; InternLM and DBRX models \cite{2023internlm,dbrx2024}; Vicuna, Alpaca, WizardLM, Tulu, Zephyr, and Starling variants \cite{vicuna2023,alpaca,xu2024wizardlm,ivison2023camels,tunstall2024zephyr,zhu2024starlingb}; and open-source models including GPT-J, GPT-NeoX, Pythia, BLOOM, OPT, GLM, T5, T0, UL2, MPT, RedPajama, YaLM, and IDEFICS \cite{gptj,black2022gptneox,biderman2023pythia,scao2022bloom,zhang2022opt,DBLP:conf/acl/DuQLDQY022,raffel2020t5,sanh2022t0,tay2023ul2,mosaicml2023mpt,redpajama,yalm,huggingface2023idefics}. The model responses are taken from HELM, VHELM, HELM Lite, and Open LLM Leaderboard v2 \cite{liang2023holistic,lee2024vhelm,liang2023helmlite,open-llm-leaderboard-v2}. When the same base model is evaluated under multiple prompting setups or leaderboard entries, we treat those variants as separate candidates in the corresponding dataset.

\section{Evaluation Metrics for LLM Outputs}
\label{sec:appendix-llm-evaluation-metrics}

All experiments use reference-based metrics that compare a model response to a reference answer. For a query $q_i$ with reference $r_i$, a task-specific metric assigns a scalar score to the response $f_j(q_i)$. Model performance on a realization is the average of these per-instance scores, and the true best model is defined with respect to this average over all instances in the realization. Higher scores are better for all metrics.

\paragraph{Exact match.}
For tasks with discrete labels or short canonical answers, we use exact match. The metric assigns a binary score indicating whether the model response is exactly equal to the reference answer. This metric is strict because a response receives credit only when it matches the reference.
\[
s_{\mathrm{EM}}(f_j(q_i), r_i) =
\indicator\{f_j(q_i) = r_i\}.
\]

\paragraph{Mathematical equivalence.}
For MATH, we use the equivalence-based scoring procedure associated with the benchmark \cite{hendrycks2021measuring}. The metric checks whether the final predicted answer is mathematically equivalent to the reference answer, rather than requiring a literal string match. This is important for mathematical reasoning tasks because the same value can be expressed in many algebraically equivalent forms.
\[
s_{\mathrm{equiv}}(f_j(q_i), r_i) =
\indicator\{f_j(q_i) \equiv r_i\},
\]
where $\equiv$ denotes mathematical equivalence.

\paragraph{F1 score.}
For open-domain and reading-comprehension question answering tasks, we use F1 score. The score is the harmonic mean of token precision and token recall between the model output and the reference answer.
\[
p_i = \frac{|T(f_j(q_i)) \cap T(r_i)|}{|T(f_j(q_i))|}, \qquad
\rho_i = \frac{|T(f_j(q_i)) \cap T(r_i)|}{|T(r_i)|},
\]
\[
s_{\mathrm{F1}}(f_j(q_i),r_i) = \frac{2p_i\rho_i}{p_i+\rho_i},
\]
where $T(\cdot)$ denotes the token multiset of a text. The score is set to zero when $p_i+\rho_i=0$.

\paragraph{BLEU.}
For machine translation, we use BLEU \cite{papineni-etal-2002-bleu}. BLEU is a family of metrics based on modified $n$-gram precision between a candidate translation and a reference, together with a brevity penalty to discourage overly short outputs. By comparing phrase overlap at multiple $n$-gram lengths, BLEU captures local fluency and phrase-level correspondence with the reference.
\[
s_{\mathrm{BLEU}\text{-}N}(f_j(q_i),r_i) = \mathrm{BP}\cdot
\exp\left(\sum_{n=1}^{N} w_n \log p_n\right),
\qquad
\mathrm{BP} = \min\left(1,\exp\left(1-\frac{|r_i|}{|f_j(q_i)|}\right)\right),
\]
where $p_n$ is the modified precision for $n$-grams, $w_n$ are the weights over $n$-gram orders, and $\mathrm{BP}$ is the brevity penalty. In our experiments, we use BLEU-4 with uniform weights over $1$-, $2$-, $3$-, and $4$-grams.

\paragraph{ROUGE.}
For summarization and captioning-style generation, we use ROUGE scores \cite{lin-2004-rouge}. ROUGE measures lexical overlap between a generated text and a reference text. In the main experiments, ROUGE-L is used. It is based on the longest common subsequence and therefore rewards in-order overlap without requiring exact contiguous matches. Additional ROUGE variants, such as ROUGE-1 and ROUGE-2, follow the same overlap-based family using unigram and bigram overlap.
\[
s_{\mathrm{ROUGE}\text{-}N}(f_j(q_i),r_i) =
\frac{\sum_{g \in \mathcal{G}_N(r_i)} \min(c_{f_j(q_i)}(g), c_{r_i}(g))}
{\sum_{g \in \mathcal{G}_N(r_i)} c_{r_i}(g)}.
\]
Here, $\mathcal{G}_N(r_i)$ is the set of reference $N$-grams and $c_{f_j(q_i)}(g)$ and $c_{r_i}(g)$ are the counts of $g$ in the generated text and reference. For ROUGE-L, the central quantity is the longest common subsequence length
\[
L_i = |\operatorname{LCS}(f_j(q_i),r_i)|,
\]
which measures how much of the reference appears in the generated text in the same order.

\paragraph{BERTScore.}
For semantic similarity in summarization, we also use BERTScore \cite{zhang2020bertscoreevaluatingtextgeneration}. BERTScore aligns tokens in the generated text and reference using contextual embeddings from a pretrained language model, then aggregates the resulting similarity scores. Compared with lexical-overlap metrics, it can assign credit to semantically similar wording even when the surface forms differ.
\[
P_B = \frac{1}{|T(f_j(q_i))|}\sum_{a \in T(f_j(q_i))}\max_{b \in T(r_i)} e(a)^\top e(b), \qquad
R_B = \frac{1}{|T(r_i)|}\sum_{b \in T(r_i)}\max_{a \in T(f_j(q_i))} e(a)^\top e(b),
\]
\[
s_{\mathrm{BERTScore}}(f_j(q_i),r_i)=\frac{2P_B R_B}{P_B+R_B}.
\]
Here, $T(\cdot)$ denotes the token sequence, $e(\cdot)$ denotes the contextual embedding of a token, $P_B$ is embedding-based precision, and $R_B$ is embedding-based recall. In our experiments, BERTScore is computed using \texttt{microsoft/deberta-large-mnli} \cite{he2021deberta}.

\section{Parameter Selection and Sensitivity}
\label{sec:appendix-temperature-selection}

The temperature parameter $\tau$ controls how sharply the posterior over candidate models responds to newly observed oracle annotations. Smaller values of $\tau$ make the posterior concentrate more strongly on models whose responses receive high scores, while larger values make posterior updates less aggressive. Thus, $\tau$ affects the update behavior of \framework, but it does not change the definition of the true best model.

As described in Section~\ref{subsec:parameter_selection}, we select $\tau$ before the final evaluation runs and without using oracle annotations from those runs. For each dataset, we first construct noisy proxy scores from pairwise similarities between model responses. We then run the same evaluation procedure over realizations using this proxy signal and select $\tau$ from the resulting identification probability curves, seeking values that reach high identification probability with fewer annotations and maintain strong average identification probability across budgets. In the main experiments, this search is performed over the grid $\{0.1, 0.2, 0.3, 0.4, 0.5, 0.6, 0.7, 0.8, 0.9, 1, 2, 3, 4, 5\}$. The selected values are reported in Section~\ref{sec:appendix-datasets-model-collections}.

The range of the grid was chosen to cover the useful regimes of the posterior update. For values below $\tau=0.1$, we observed that the posterior can become concentrated on a single model after only a few annotations. This makes the method commit very early based on limited evidence, which is undesirable when the goal is to identify the best model reliably. At the other end, values above $\tau=5.0$ make the update sufficiently flat that additional increases produce only very small changes in the posterior and in the selected queries. The grid therefore covers the region where changes in $\tau$ have a visible effect on the behavior of \framework.

The selected temperatures vary across datasets. We observe that metrics such as ROUGE, BLEU, and BERTScore often produce relatively small numerical differences between strong models, and in several of these settings the noisy proxy selected smaller temperatures. Exact match and equivalence metrics can produce sharper score gaps, and some of these datasets selected larger temperatures. This pattern should be interpreted as an empirical tendency rather than a strict rule. The best value of $\tau$ also depends on the separation between candidate models, the difficulty of the dataset, and the distribution of model responses.

\begin{figure}[t]
  \centering
  \includegraphics[width=\textwidth]{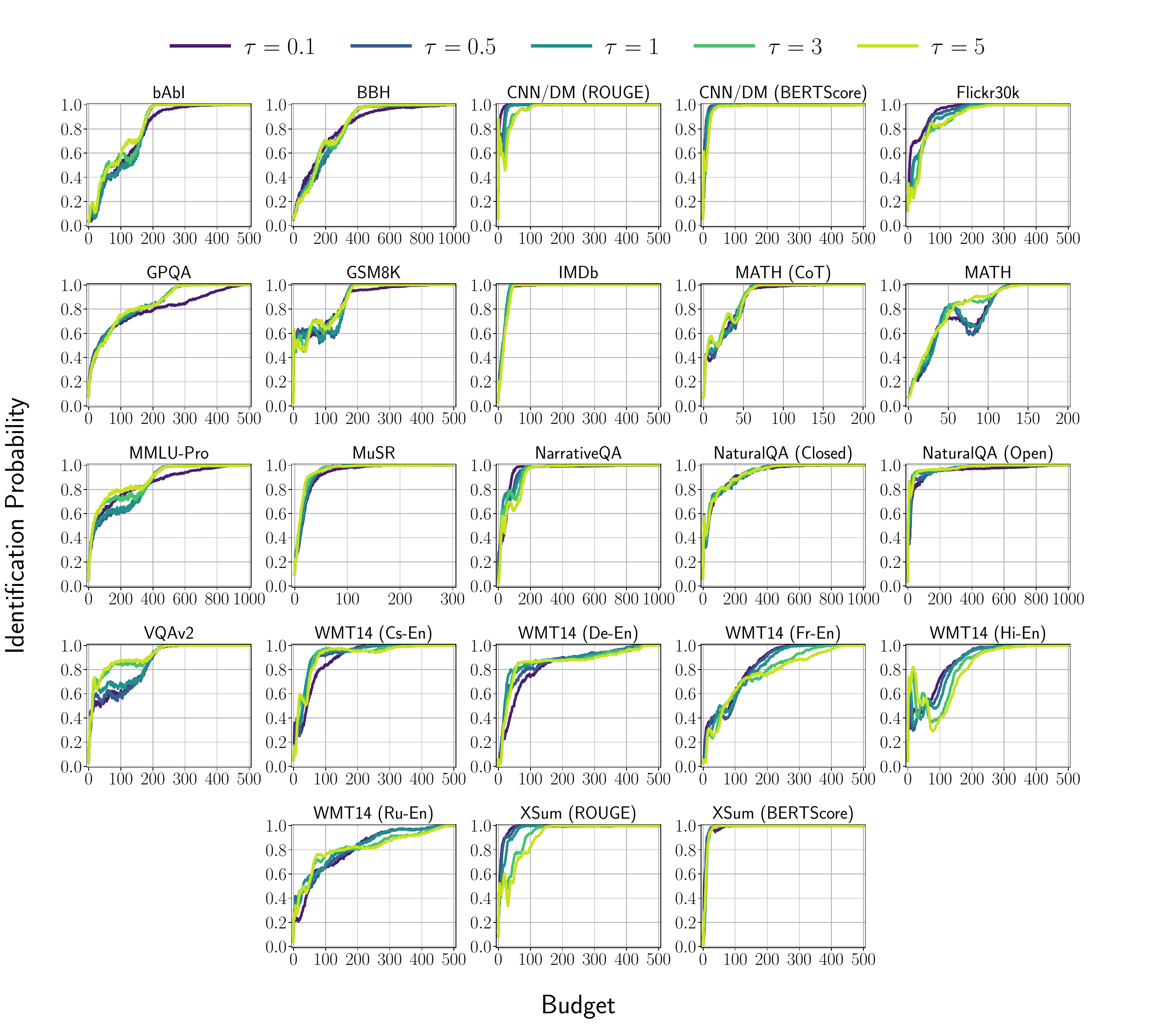}
  \caption{Sensitivity of \framework\ to the temperature parameter $\tau$ across the $23$ datasets.}
  \label{fig:parameter_sensitivity}
\end{figure}

Figure~\ref{fig:parameter_sensitivity} reports the sensitivity of \framework\ to the temperature parameter using the shared grid $\{0.1, 0.5, 1.0, 3.0, 5.0\}$ across all $23$ datasets. The results show that \framework\ maintains high annotation efficiency across a broad range of temperature values. At the same time, $\tau$ can affect early-budget behavior and annotation efficiency on some datasets, including summarization, translation, visual question answering, and the newer reasoning-heavy tasks. Consistent with the discussion above, these differences appear partly related to the scale and discreteness of the evaluation metric: some datasets favor lower temperatures, while others favor higher temperatures. However, once $\tau$ is in a suitable region for the dataset, further fine-grained tuning typically changes the curves only modestly. This supports the use of the noisy proxy selection procedure as an annotation-free calibration step, while also indicating that performance is not dependent on a single precisely tuned value.

In practice, we recommend using the noisy proxy procedure to choose $\tau$ for the target dataset. The sensitivity results suggest that the choice should be calibrated, but does not require fine-grained tuning, since nearby reasonable values often produce similar behavior.

\section{Computational Complexity}
\label{sec:appendix-computational-complexity}

In addition to its strong performance, \framework\ is computationally lightweight. All model selection experiments were run on a computer with a 13th Gen Intel Core i7-1355U CPU (12 logical threads) and 16 GB of RAM. The experiments use precomputed model outputs and pairwise response similarities, so the selection procedure does not require any LLM inference, fine-tuning, or gradient computation.

At step $t$, \framework\ scores each remaining unlabeled query by computing the posterior-weighted pairwise similarity over all model pairs, which costs $O((n-t)m^2)$, and then updates the posterior over models in $O(m)$. The total cost of one realization is therefore
$O\!\left(m^2 \sum_{t=0}^{b-1} (n-t) + bm\right)
=
O\!\left(\frac{m^2 b(2n-b+1)}{2} + bm\right)$,
which is $O(n^2m^2)$ when $b=n$. The memory cost is dominated by the precomputed pairwise similarity tensor, which requires $O(nm^2)$ space for a realization, together with $O(nm)$ space for oracle scores and $O(m)$ space for the model posterior. This makes \framework\ more computationally involved than non-adaptive baselines such as \textsc{Random}, \textsc{Margin}, and \textsc{Min Agreement}, but lighter in its dependence on the number of models than the risk-based \textsc{VMA} and \textsc{AMC} baselines, whose implementations scale cubically in $m$.

For the complete final method comparison runs on each dataset, including \framework\ and all baselines over $1000$ realizations, the implementation was parallelized over independent realizations. These runs took between $21.7$ seconds for MuSR and $399.2$ seconds ($6.7$ minutes) for Natural Questions (Open-Book), with a mean of $117.5$ seconds per dataset. Across all $23$ datasets, the total wall-clock runtime was $45.1$ minutes.

\section{Existing Assets, Licenses, and Terms}
\label{sec:appendix-assets-licenses}

Our experiments use existing benchmark datasets, model output collections, and standard text evaluation metrics. We cite the original creators of these assets in Section~\ref{subsec:dataset_model_collections} and Appendix~\ref{sec:appendix-datasets-model-collections}. The main model output and benchmark result sources used in this work are HELM~\cite{liang2023holistic}, VHELM~\cite{lee2024vhelm}, HELM Lite~\cite{liang2023helmlite}, and Open LLM Leaderboard v2~\cite{open-llm-leaderboard-v2}. HELM, VHELM, and HELM Lite are distributed through the HELM ecosystem, whose framework code is released under the Apache License 2.0; benchmark assets and model outputs follow the licenses and terms of the underlying datasets and model providers. Open LLM Leaderboard v2 is distributed through Hugging Face, includes model-specific Hub license metadata, and is used according to the Hugging Face terms and the corresponding model and benchmark terms.

The benchmark datasets used in our experiments are used under their respective licenses and terms:
\begin{itemize}
    \item bAbI~\cite{weston2015towards}: BSD License for the bAbI task repository.
    \item BBH~\cite{suzgun2022challenging}: MIT License.
    \item GPQA~\cite{rein2023gpqa}: Creative Commons Attribution 4.0 International (CC BY 4.0).
    \item MMLU-Pro~\cite{wang2024mmlupro}: MIT License.
    \item MuSR~\cite{sprague2023musr}: MIT License.
    \item CNN/DailyMail~\cite{hermann2015teaching}: Apache License 2.0 for the commonly used dataset release.
    \item XSum~\cite{narayan-etal-2018-dont}: source terms; common redistributed versions report either CC BY-SA 4.0 or an unspecified license, so we follow the terms of the source used by the corresponding benchmark collection.
    \item Flickr30k~\cite{flickr30k}: source terms for Flickr30k annotations and Flickr Terms of Use for the underlying images.
    \item GSM8K~\cite{cobbe2021gsm8k}: MIT License.
    \item MATH~\cite{hendrycks2021measuring}: MIT License.
    \item IMDb~\cite{maas-etal-2011-learning}: IMDb non-commercial dataset terms / Large Movie Review Dataset source terms.
    \item NarrativeQA~\cite{kocisky-etal-2018-narrativeqa}: Apache License 2.0.
    \item Natural Questions~\cite{kwiatkowski2019natural}: Creative Commons Share-Alike 3.0.
    \item VQAv2~\cite{goyal2017making}: Creative Commons Attribution 4.0 International (CC BY 4.0) for the annotations, with images governed by their underlying source terms.
    \item WMT14~\cite{bojar-etal-2014-findings}: WMT14/statmt source terms, with terms depending on the underlying language-pair resources.
\end{itemize}

We do not release any new model weights. The released artifacts are intended to support reproducibility of the active model selection experiments and include code, configuration files, and processed experimental data where redistribution is permitted. For assets that cannot be redistributed directly, we provide instructions for obtaining them from their original sources.

\end{document}

%% file: math_commands.tex
\usepackage{amsmath,amsfonts,bm}

\newcommand{\framework}{\textsc{Select-LLM}}

\def\eqref#1{equation~\ref{#1}}

\def\Eqref#1{Equation~\ref{#1}}

\def\1{\bm{1}}

\DeclareMathAlphabet{\mathsfit}{\encodingdefault}{\sfdefault}{m}{sl}
\SetMathAlphabet{\mathsfit}{bold}{\encodingdefault}{\sfdefault}{bx}{n}

\DeclareMathOperator*{\argmax}{arg\,max}
\DeclareMathOperator*{\argmin}{arg\,min}